\def\year{2020}\relax
\documentclass[letterpaper]{article} % DO NOT CHANGE THIS
\usepackage{aaai20}  % DO NOT CHANGE THIS
\usepackage{times}  % DO NOT CHANGE THIS
\usepackage{helvet} % DO NOT CHANGE THIS
\usepackage{courier}  % DO NOT CHANGE THIS
\usepackage[hyphens]{url}  % DO NOT CHANGE THIS
\usepackage{graphicx} % DO NOT CHANGE THIS
\usepackage{graphicx}  % DO NOT CHANGE THIS
\usepackage{subcaption}
\usepackage{tabu}
\usepackage{multirow}
\newcommand\xrowht[2][0]{\addstackgap[.5\dimexpr#2\relax]{\vphantom{#1}}}
\usepackage{algorithm}
\usepackage{algorithmic}
\usepackage{pdfpages}

\usepackage{todonotes}

\usepackage{amsmath,amsfonts,bm}
\usepackage{mathtools, cuted}
\def\rvx{{\mathbf{x}}}
\def\rvy{{\mathbf{y}}}
\def\rvz{{\mathbf{z}}}

\def\vx{{\bm{x}}}

\DeclareMathOperator*{\argmax}{arg\,max}

\newcommand{\E}{\mathbb{E}}
\usepackage{stackengine}
\def\delequal{\mathrel{\ensurestackMath{\stackon[1pt]{=}{\scriptscriptstyle\Delta}}}} % define triangle equality
\newcommand{\KL}{D_{\mathrm{KL}}}
\usepackage{tikz}
\newcommand*\circled[1]{\tikz[baseline=(char.base)]{
            \node[shape=circle,draw,inner sep=2pt] (char) {#1};}}
\title{Balancing Reconstruction Quality and Regularisation in Evidence Lower Bound for Variational Autoencoders}
\author{Shuyu Lin\textsuperscript{\rm 1}, Stephen Roberts\textsuperscript{\rm 1}, Niki Trigoni\textsuperscript{\rm 1}, Ronald Clark\textsuperscript{\rm 2}\\ 
}
\begin{document}

\maketitle

\begin{abstract}
A trade-off exists between reconstruction quality and the prior regularisation in the Evidence Lower Bound (ELBO) loss that Variational Autoencoder (VAE) models use for learning. There are few satisfactory approaches to deal with a balance between the prior and reconstruction objective, with most methods dealing with this problem through heuristics. 
In this paper, we show that the noise variance (often set as a fixed value) in the Gaussian likelihood $p(\rvx|\rvz)$ for real-valued data can naturally act to provide such a balance.
By learning this noise variance so as to maximise the ELBO loss, we automatically obtain an optimal trade-off between the reconstruction error and the prior constraint on the posteriors. This variance can be interpreted intuitively as the necessary noise level for the current model to be the best explanation of the observed dataset. Further, by allowing the variance inference to be more flexible it can conveniently be used as an uncertainty estimator for reconstructed or generated samples.
% conditioned on specific latent encodings and to be variable across different data dimensions, 
We demonstrate that optimising the noise variance is a crucial component of VAE learning, and showcase the performance on MNIST, Fashion MNIST and CelebA datasets. We find our approach can significantly improve the quality of generated samples whilst maintaining a smooth latent-space manifold to represent the data. The method also offers an indication of uncertainty in the final generative model.
\end{abstract}

\section{Introduction}
Variational Auto-encoders (VAEs, \cite{VAE,diff-VAE}) are a stable and efficient approach to unsupervised learning.
VAEs naturally combine two learning outcomes: enabling generation of data similar to the observations and secondly offering a probabilistic embedding scheme, in which the statistics at both sample and group levels satisfy required constraints. VAEs can, for example, be used for dimensionality reduction, feature extraction and efficient anomaly detection in the latent space.

Learning with a VAE naturally requires a balance between reconstruction loss minimisation and prior constraint enforcement. Several papers discuss the importance of this balance and the impact of different trade-off settings on the learned VAE models \cite{fixing-ELBO,disentangling-disentanglement}. 
In \cite{beta-VAE}, the authors propose scaling up the prior regularisation term in an attempt to promote disentanglement in the learnt representation. On the other hand, in \cite{deep-info-bottleneck} more emphasis is placed on the reconstruction loss term, to promote maximal information retention in the learned representation.
The trade is most commonly adapted on a problem by problem basis \cite{visually-grounded-imagination,JMVAE,SCAN-b-VAE}.

In this paper, we propose a mechanism to optimise the trade-off between prior and reconstruction loss. We show that, by parameterising this balance via a variance parameter, we can achieve significant performance gains, both in terms of achieving a closer bound to the true data log likelihood as well as improving the generated sample quality. Unlike the hyperparameter introduced in \cite{beta-VAE}, this variance hyperparameter intuitively represents the noise level in the dataset conditioned on the model explaining the observations. Furthermore, we develop a robust algorithm to learn to predict the variance hyperparameter conditioned on a given input. We show that the predicted variance can be used as an effective uncertainty estimator for reconstructed or generated samples. 

By altering the strength of the prior regularisation term, the gap between the marginal latent distribution and the prior, measured for example using the Kullback Leibler (KL) divergence between the aggegate posterior and prior of the learnt model, becomes more prominent. This can significantly damage the quality of the generated samples as measured by common perceptual metrics such as the Inception Score (IS) \cite{salimans2016improved} or the Frechet Inception Distance (FID) \cite{heusel2017gans}. To avoid this problem, we propose a simple approximation to the learned latent marginal distribution and use this approximate distribution, rather than the prior, to generate samples. 

In summary, our contributions are:
\begin{itemize}
    \item We show how optimisation of the variance hyperparameter under the VAE ELBO loss allows us to automate the trade-off between the reconstruction loss and the prior regularisation and we propose a stable learning procedure to accomplish this optimisation.
    %\item We provide an effective uncertainty estimator for VAE models. 
    \item We study the impact of sampling from the learned latent marginal distribution as opposed to the prior and propose to use an approximate marginal distribution instead of the prior to generate samples. 
\end{itemize}

\section{Background}
\subsection{VAE ELBO Loss}
Given a dataset of $N$ observations $\mathcal{D}_N = \{\vx_1, \vx_2, \cdots, \vx_N\}$ and a latent variable model $p_{\theta}(\rvx, \rvz) = p_{\theta}(\rvx | \rvz)p(\rvz)$, the learning objective in VAEs maximises the marginal likelihood for all the data points in $\mathcal{D}_N$ under the model parameter $\theta$, i.e. 
\begin{equation} \label{eq:VAE-initial-obj}
    \argmax_{\theta} \E_{p_{\mathcal{D}}(\rvx)}
    [\log p_{\theta}(\rvx)],  
\end{equation}
where $p_{\mathcal{D}}$ describes the true data distribution and $p_{\theta}(\rvx)=\int p_{\theta}(\rvx|\rvz)p(\rvz) \textrm{d}\rvz$ is the marginal distribution of $\rvx$ under the latent variable model by integrating out $\rvz$ according to a prior distribution $p(\rvz)$. Directly evaluating Equation (\ref{eq:VAE-initial-obj}) is often not feasible, because $p_{\theta}(\rvx|\rvz)$ is often parameterised by a neural network and integration over such a function cannot be easily evaluated. 
Instead, we can use variational evidence lower bound (ELBO) $\mathcal{L}(\rvx; \theta, \phi)$ on $\log p_{\theta}(\rvx)$ as an alternative objective by introducing an amortized inference model $q_{\phi}(\rvz|\rvx)$, as shown below:
\begin{align}
    &\log p_{\theta}(\rvx) 
    \geq \, \mathcal{L}(\rvx; \theta, \phi) \\
    &\delequal \;
    \log \;p_{\theta}(\rvx)
     - 
    \KL\big[\,q_{\phi}(\rvz|\rvx) \Vert  p_{\theta}(\rvz|\rvx)\big] \\
    &\delequal \; 
    \E_{q_{\phi}(\rvz|\rvx)}\big[\log p_{\theta}(\rvx|\rvz)\big]
    -
    \KL\big[\,q_{\phi}(\rvz|\rvx) \Vert p(\rvz)\big]. \label{eq:ELBO-def}
\end{align}
Substituting $\mathcal{L}(\rvx; \theta, \phi)$ in (\ref{eq:ELBO-def}) into the VAE objective (\ref{eq:VAE-initial-obj}) and evaluating the expectation wrt $p_{\mathcal{D}}$ with an empirical approximation, i.e. $p_{\mathcal{D}} (\rvx) \approx \frac{1}{N} \sum_{i=1}^N \delta(\vx_i)$,
we have the final VAE learning objective:
\begin{align}
    % \E_{p_{\mathcal{D}}(\rvx)}[\log p_{\theta}(\rvx)]
    % \;\geq\;
    % \E_{p_{\mathcal{D}}(\rvx)}[\mathcal{L}(\rvx; \theta, \phi)]
    % % &\;\approx\;
    % % \frac{1}{N}\sum_{i=1}^{N} \mathcal{L}(\vx_i; \theta, \phi), 
    % % \label{eq:ELBO-monte-carlo}\\
    % % \textrm{where} \,\, \mathcal{L}(\vx_i; \theta, \phi) =
    % \;=\;
    \frac{1}{N}\sum_{i=1}^{N}  \Big[
    \underbrace{\E_{q_{\phi}(\rvz|\vx_i)}\big[\log p_{\theta}(\vx_i|\rvz)\big]}_\text{\circled{1} \textrm{Reconstruction likelihood}}
    -
    \underbrace{\KL\big[\,q_{\phi}(\rvz|\vx_i) \,\Vert\, p(\rvz)\big]}_\text{\circled{2} \textrm{Prior constraint}} \Big]. \label{eq:VAE-ELBO-obj}
\end{align}
Facilitated with the reparameterisation trick \cite{repara-trick}, the above ELBO loss can be optimised with stochastic gradient descent algorithms, leading to an efficient learning method \shortcite{VAE,diff-VAE}. 

\subsection{Competing Learning Objectives in ELBO}
The two terms in the ELBO loss (Eq. \ref{eq:VAE-ELBO-obj}) control different behaviours of the model. The \circled{\small{1}} \emph{reconstruction likelihood} term attempts to reconstruct the input $\vx_i$ as faithfully as possible. The \circled{\small{2}} \emph{prior constraint} term enforces the posterior distribution of $\vx_i$ under the encoding function to comply with an assumed prior $p(\rvz)$ and if the constraint is effectively imposed, the aggregate posterior of the entire dataset
\cite{VampPrior} is likely to be very close to the prior, indicating the data distribution has been successfully projected to the target prior in the latent space. Optimizing the two losses together, we would hope to simultaneously achieve the best performance in both terms.

Unfortunately, the two losses are often conflicting. To see this, consider if \circled{\small{1}} is maximized, then the embeddings of different input samples $\vx_i$ and $\vx_j$ can be easily distinguished, i.e. $\KL \big[q_{\phi}(\rvz|\vx_i) \,\Vert\, q_{\phi}(\rvz|\vx_j) \big]$ should be large. On the other hand, if \circled{\small{2}} is satisfied, then all sample embeddings should approach $q(\rvz)$, i.e. $\KL \big[q_{\phi}(\rvz|\vx_i) \,\Vert\, q_{\phi}(\rvz|\vx_j) \big]$ should be zero. Therefore,  \circled{\small{1}} and \circled{\small{2}} almost always influence each other.
% there is a constant competition between the two losses in the ELBO.
As a result, finding the optimal balance between the two terms is a crucial part in the VAE learning process.
Unfortunately, there has not been a satisfying solution and most methods handle this problem by introducing weight hyperparameters on the two losses and choose their values based on heuristics \shortcite{beta-VAE,beta-VAE2,deep-info-bottleneck}.  

% $\beta$-VAE proposes to introduce a hyperparameter $\beta$ in Equation \ref{eq:ELBO-per-sample} and uses this $\beta$ to control the relative weights between the 2 terms in the objective function. However, such manipulation modifies the ELBO objective and the resulted objective function no longer satisfies a lower bound of actual learning outcome, i.e. the marginal log likelihood $\E_{p_{\mathcal{D}}(\rvx)} [\log p_{\theta}(\rvx)]$. This means the optimised parameters under $\beta$-VAE objective might not be the parameters that explain the dataset the best. 

\section{Our Proposal} \label{Sec:Our-method}
In this work, we propose a method that automatically finds the best balance between the two contradicting objectives in the VAE ELBO loss.
\subsection{Gaussian Likelihood in the Generative Model}
For real-valued data $\rvx$, we can use a Gaussian distribution to model the conditional distribution $p_{\theta}(\vx_i|\rvz)$ in the generative model, i.e. 
\begin{equation} \label{eq:conditional-density}
    p_{\theta}(\vx_i^k|\rvz) = \frac{1}{{(2\pi)}^{\frac{1}{2}} \sigma} 
        \exp{\Bigg(-\frac{\big(g_{\theta}^k(\rvz) - \vx_i^k\big)^2}{2 \sigma^2}\Bigg)},
\end{equation}
where $\sigma^2$ is a common global variance parameter, reflecting the global noise properties of the data; $g_{\theta}(\cdot)$ represents a nonlinear mapping that transforms an encoding $\rvz$ in the latent space to the data space; and $k$ represents the $k$-th dimension of the data variable $\rvx$. If we assume that different dimensions in $\rvx$ are independent, then the conditional density of the complete data variable is
\begin{equation}
    p_{\theta}(\vx_i|\rvz) = \prod_{k=1}^{d} p_{\theta}(\vx_i^k|\rvz),
    % = \frac{1}{{(2\pi)}^{\frac{d}{2}} {\sigma}^d} 
    % \exp{\Bigg(-\frac{1}{2 \sigma^2} \sum_{k=1}^{d} \big(g_{\theta}^k(\rvz) - \vx_i^k\big)^2 \Bigg)},
\end{equation}
where $d$ is the dimension of the data variable $\rvx$.
Therefore, $\log p_{\theta}(\vx_i|\rvz)$ in the ELBO loss given by (\ref{eq:VAE-ELBO-obj}) can be computed as
\begin{equation} \label{eq:log-recons-likelihood}
    \log p_{\theta}(\vx_i|\rvz)
    =
    -\frac{d}{2}\log(2\pi)
    -d\log \sigma
    -\frac{1}{2\sigma^2}\sum_{k=1}^{d}\big(g_{\theta}^k(\rvz) - \vx_i^k\big)^2.
\end{equation}
Notice the terms inside the summation are element-wise square errors between the generated sample and the original sample and $\frac{1}{2\sigma^2}$ naturally appears as a weighting factor on the sum of square error term. When $\log p_{\theta}(\vx_i|\rvz)$ is maximised, the $\log \sigma$ term has an important regularizing effect on the $\sigma$ value: it prevents $\sigma$ from taking very large values, which would allow the generator function $g(\cdot)$ to produce arbitrarily bad reconstruction.

\subsection{ELBO Loss with A Global Variance Parameter}
Substituting (\ref{eq:log-recons-likelihood}) into (\ref{eq:VAE-ELBO-obj}), we have the overall ELBO loss with the Gaussian likelihood assumption as
\begin{equation} \label{eq:ELBO-with-sigma}
    \begin{split}
    &\E_{p_{\mathcal{D}}(\rvx)}[\log p_{\theta}(\rvx)]
    \geq 
    -\frac{1}{N}\sum_{i=1}^{N}  \Bigg[
    \frac{d}{2}\log(2\pi)
    +d\log \sigma 
    + \\
    &\underbrace{\frac{1}{2\sigma^2}}_\text{\circled{3}} 
    \E_{q_{\phi}(\rvz|\vx_i)} \bigg[ \underbrace{\sum_{k=1}^{d}\big(g_{\theta}^k(\rvz) - \vx_i^k\big)^2}_\text{\circled{1}} \bigg] 
    +\underbrace{\KL\Big[\,q_{\phi}(\rvz|\vx_i) \,\Vert\, p(\rvz)\Big]}_\text{\circled{2}}
    \Bigg].
    \end{split}
\end{equation}
From Equation (\ref{eq:ELBO-with-sigma}), we can see that term \circled{\small{3}} represents a \emph{relative weighting} $\frac{1}{2\sigma^2}$ between the two competing learning objectives; \circled{\small{1}} the reconstruction error; and \circled{\small{2}} prior constraint on the posteriors $q_{\phi}(\rvz|\vx_i)$. If we optimise the ELBO in (\ref{eq:ELBO-with-sigma}) w.r.t. $\theta$, $\phi$ and $\sigma$, then we will reach a maximised lower bound of the data likelihood where the model automatically balances between the objectives of minimizing the information loss through the auto-encoding process while still having the latent marginal distribution remain close to the prior distribution. The optimal $\sigma^2$, which is obtained at the maximal ELBO and denoted as $\sigma_{*}^2$, can be interpreted as the amount of noise that has to be assumed in the dataset for the current reconstruction to be considered as the best explanation of the observed samples.

\subsection{Closed-form Solution for the Variance} \label{Sec:closed-form-solution-sigma}
For fixed $\theta$ and $\phi$, we can derive a closed form solution for the optimal global variance $\sigma_{*}^2$. To compute $\sigma_{*}^2$, we take derivative of $\E_{p_{\mathcal{D}}(\rvx)}[\log p_{\theta}(\rvx)]$ in Equation (\ref{eq:ELBO-with-sigma}) wrt $\sigma$ and set it to zero. Then
\begin{equation} \label{eq:global-sigma-closed-form}
    {\sigma_{*}}^2 = 
     \frac{1}{d}\E_{q_{\phi}(\rvz|\vx_i)} \bigg[\sum_{k=1}^{d} \big(g_{\theta}^k(\rvz) - \vx_i^k\big)^2 \bigg].
\end{equation}
The significance of this result is that it improves the stability of the learning process with the additional $\sigma^2$ parameter, as we can always find a local minimum of the ELBO loss by updating $\sigma^2$ and the model parameters $\theta$ and $\phi$ iteratively.

\subsection{More Flexible Variance Estimation} \label{Sec:pixel-sigma}
So far we have taken $\sigma^2$ to be the same value for the entire dataset and across all data dimensions, but nothing stops us here. We can extend the $\sigma^2$ estimation to be conditioned on particular encodings, be that an encoding of a real data sample or a sampled latent code from the prior distribution. We can also allow $\sigma^2$ to differ across data dimensions and such $\sigma^2$ can conveniently highlights the regions with high uncertainty in a reconstructed or generated sample. 
To achieve these goals, we need to replace the global variance $\sigma^2$ in Equation (\ref{eq:conditional-density}) by a variance estimation function $\sigma_{\theta}^{k}(\rvz)$ which is parameterised by $\theta$ and conditioned on data dimension $k$ and an input encoding $\rvz$. The corresponding $\log p_{\theta}(\vx_i|\rvz)$ now becomes
\begin{equation} \label{eq:log-recons-likelihood-pixel-sigma}
    -\frac{d}{2}\log(2\pi)
    -\sum_{k=1}^{d}\bigg[\frac{1}{2\big(\sigma_{\theta}^{k}(\rvz)\big)^2} 
        \big(g_{\theta}^k(\rvz) - \vx_i^k\big)^2 
    +\log \sigma_{\theta}^{k}(\rvz) \bigg].
\end{equation}
Substituting (\ref{eq:log-recons-likelihood-pixel-sigma}) in (\ref{eq:VAE-ELBO-obj}), we get the ELBO loss associated with the input dependent variance estimation and learning can be done by optimising this ELBO wrt to all model parameters $\theta$ and $\phi$. 

However, we notice that learning to predict $\sigma_{\theta}^{k}(\rvz)$ together with learning the parameters for the auto-encoding task from scratch turns out to be extremely challenging and the optimisation often gets stuck at predicting very small variance values. This is because the variance prediction is highly dependent on the reconstruction error given by the current model. At early stage when both the encoder and decoder are inaccurate, the reconstruction error is large and has high variance. This causes random updates in the variance prediction module. If the variance happens to arrive at a small value where the gradient of the $\log \sigma_{\theta}^{k}(\rvz)$ term dominates, then it is very hard for the gradient update to escape the strong negative gradient from the $\log$ function. To prevent this happening and obtain stable learning, we propose a staged learning process where we start with learning the global variance parameter until certain condition of convergence is reached and then we switch to the input dependent variance prediction module with the model parameters ($\phi$ and $\theta$) continuing updates from the optimised values given by the previous stage. We also use the optimised global variance as an effective lower bound to prevent the predicted variance from getting stuck at small values. The learning procedure is summarised in Algorithm \ref{alg:alpha-VAE}. 

% At test time, we use the optimised $\sigma_{\theta}(\rvz)$ to estimate the uncertainty of a generated data sample from the encoding $\rvz$. Regions with high $\sigma$ values indicate a pattern that is rare in the training dataset and hence the model is less confident about. 

\subsection{Importance of Aggregate Posterior} \label{sec:GM-agg-posterior}

Even when optimising the variance hyperparameter, the learnt posteriors might be quite different from the assumed prior distribution. If this is the case, then using the prior distribution to generate samples may no longer be a good idea. A better alternative is to generate samples from the aggregate posterior $q_{\phi}(\rvz)$, which can be considered as the data distribution in the latent space, defined \shortcite{VampPrior} as 
\begin{equation} \label{eq:agg-posterior}
    q_{\phi}(\rvz)
    % =
    % \int q_{\phi}(\rvz|\rvx) p_{\mathcal{D}}(\rvx) \textrm{d}\rvx
    =
    \E_{p_{\mathcal{D}}(\rvx)} [q_{\phi}(\rvz|\rvx)]
    \approx
    \frac{1}{N} \sum_{i=1}^N q_{\phi}(\rvz|\vx_i).
\end{equation}
The approximation in (\ref{eq:agg-posterior}) is made by evaluating the expectation w.r.t $p_{\mathcal{D}}(\rvx)$ empirically for $\mathcal{D}_N$ and with the isotropic Gaussian assumption for posteriors $q_{\phi}(\rvz|\rvx)$, the aggregate posterior is effectively a mixture of Gaussian distribution with $N$ mixture components. However, there are two problem with generating data samples directly from the aggregate posterior: 1) large memory is needed to store the statistics of all $N$ components in the aggregate posterior distribution and 2) as training data is finite, the aggregate posterior is likely to be overfitted to the training samples, which means that only samples very similar to the observed data will be generated. 

We propose to address both problems by using a simpler approximate distribution to the true aggregate posterior. There are many choices for the approximate distribution. In our case, we simply use a Gaussian mixture distribution $\hat{q}(\rvz)$ with $M$ components ($M \ll N$ and we often take $M$=30 for simple datasets, such as MNIST and Fashion MNIST), as shown below: 
\begin{equation}
    q_{\phi}(\rvz) \approx
    \hat{q}(\rvz) = 
    \sum_{m=1}^{M}w_m \mathcal{N}(\rvz; \mu_m, \Sigma_m), 
\end{equation}
where $w_m$ is the weight for each Gaussian mixture ($w_m > 0$ and $\sum_m w_m =1$) and $\mu_m$ and $\Sigma_m$ are the mean and covariance matrix for the $m$-th Gaussian mixture.
The Gaussian mixture distribution is a good choice for the following reasons: 1) it is sufficiently expressive to represent the major modes and the low density regions that $q_{\phi}(\rvz)$ might possess; 2) with limited number of components, it is much simpler than the original aggregate posterior, significantly reducing the risk of overfitting; and 3) there are efficient algorithms, such as EM \cite{PRML} and Variational Bayesian inference methods \cite{variational-bayes-GMM}, to derive the Gaussian mixture approximation.  

\begin{algorithm}[!ht]
   \caption{VAE learning with variance prediction}
   \label{alg:alpha-VAE}
\begin{algorithmic}
   \STATE $\phi$, $\theta$ $\leftarrow$ Initialize parameters
   \STATE $\sigma$ $\leftarrow$ Initialize as 1
   \STATE $\epsilon_1$, $\epsilon_2$, n\_epoch\_1, n\_epoch\_2 $\leftarrow$ Set stopping condition
   \STATE Set epoch counter $e = 0$ 
   \WHILE{e $<$ n\_epoch\_1 or $\delta_{\text{ELBO}} < \epsilon_1$}
   \STATE Update $\phi$, $\theta$ and $\sigma$ using AdamOptimizer on the ELBO loss defined in Equation (\ref{eq:ELBO-with-sigma})
   \STATE $\delta_{\text{ELBO}} = \textrm{ELBO}_{e} - \textrm{ELBO}_{e-1}$
   \STATE e = e + 1
   \ENDWHILE
   \WHILE{e $<$ n\_epoch\_2 or $\delta_{\text{ELBO}} < \epsilon_2$}
   \STATE Freeze $\sigma^2$ value as $\sigma_{\textrm{dataset}}^2$
   \STATE Add the input dependent variance module with $\big(\sigma_{\theta}^{k}(\rvz)\big)^2 = v_{\theta}^{k}(\rvz) + \frac{1}{2}\sigma_{\textrm{dataset}}^2$, \footnote{where $v_{\theta}^{k}(\rvz)$ indicates a function that takes $\rvz$ as input and predicts the $k$-th dimension variance}
   \STATE Update $\phi$, $\theta$ using AdamOptimizer on the ELBO loss with the $\log p_{\theta}(\vx_i|\rvz)$ term given by Equation (\ref{eq:log-recons-likelihood-pixel-sigma}).
   \STATE $\delta_{\text{ELBO}} = \textrm{ELBO}_{e} - \textrm{ELBO}_{e-1}$
   \STATE e = e + 1
   \ENDWHILE
   \STATE \textbf{return} $\phi, \theta, \sigma_{\textrm{dataset}}^2$
\end{algorithmic}
\end{algorithm}

\section{Related Work}
\textbf{Optimising the variance parameter in a likelihood model} has been adopted in Mixture Density Networks \cite{MDN} and Gaussian process regression models \cite{GPML}. Both works aim to learn a model for a regression task and use an isotropic Gaussian likelihood to model the conditional distribution $p(\rvy|\rvx)$ between an input $\rvx$ and an output $\rvy$. Although the nature of their learning tasks is vastly different from ours (supervised vs unsupervised), they emphasise the same message as our proposal that learning the variance parameter in the likelihood model for noisy observations is an integral part of the learning process and should not be omitted. Two major differences between our proposal and their treatment to the variance parameter are: 1) we extend the variance prediction to more flexible settings, such as being conditioned on an encoding and variable across data dimensions, whereas they only use a single-valued variance parameter; 2) the learning objective in their models only contains the Gaussian likelihood term and therefore they do not have the tradeoff between reconstruction and regularisation that we are faced with when using the ELBO loss. In the original VAE paper \cite{VAE}, estimation of a similar variance in the decoder is briefly mentioned in the appendix, where their decoder estimates a single-valued variance for a given input sample. However, such prediction is often unstable and thus our proposal uses a two-stage training procedure to stabilise the training.  

\textbf{The undesirable gap between the aggregate posterior and prior} has been noted in \shortcite{VAE-suboptimality}. Many works have been proposed to mitigate such a gap by using either a more expressive model for sample posteriors $q_{\phi}(\rvz|\vx_i)$ \cite{expressive-post-1,expressive-post-2,expressive-post-3,expressive-post-4} or a more flexible model for the prior $p(z)$ \cite{VampPrior,GMM-prior}. Our treatment of the gap differs from all the aforementioned works, as we realise that such gap is hard to be eliminated. Hence, the best remedy is to replace the prior with a distribution $\hat{p}(\rvz)$ that is better matched with the amortized posterior $q_{\phi}(\rvz|\rvx)$ and the generative likelihood $p_{\theta}(\rvx|\rvz)$ so that the gap between the amortized posterior $q_{\phi}(\rvz|\rvx)$ and the true posterior from the generative model $p_{\theta}(\rvz|\rvx)=\frac{p_{\theta}(\rvx|\rvz)\hat{p}(\rvz)}{\int p_{\theta}(\rvx|\rvz)\hat{p}(\rvz) \textrm{d}\rvz}$ can be effectively reduced, giving a closer bound to $\log p(\rvx)$. 

\section{Experiment Results}
\newcolumntype{A}{ >{\centering\arraybackslash} m{2.3cm} }
\newcolumntype{B}{ >{\centering\arraybackslash} m{2.5cm} }
% \begin{center}
\begin{table*}[t]
    \centering
    \caption{Comparison of the aggregate posterior and its Gaussian mixture approximation (32 mixtures) for models learnt under different $\sigma^2$ values for MNIST dataset. $\KL(\hat{q}(\rvz) || p(\rvz))$ denotes the gap between the approximate aggregate posterior and the prior distribution and is evaluated by Monte Carlo estimation of 10k samples (10 runs).}
    \begin{tabular}{ |B|AAAA|A| } 
        % \xrowht{10pt}
        \hline
        $\sigma^2$ & 1 & 0.5 & \textbf{0.035} (Optimal) & 0.01 & Prior \\[0.4ex]\hline
        Aggregate posterior %$q_{\phi}(\rvz)$ 
        & \includegraphics[width=2cm, height=2cm]{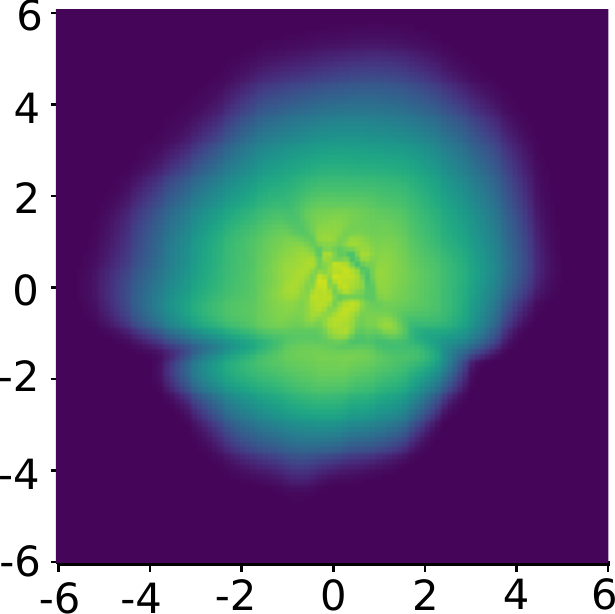}
        & \includegraphics[width=2cm, height=2cm]{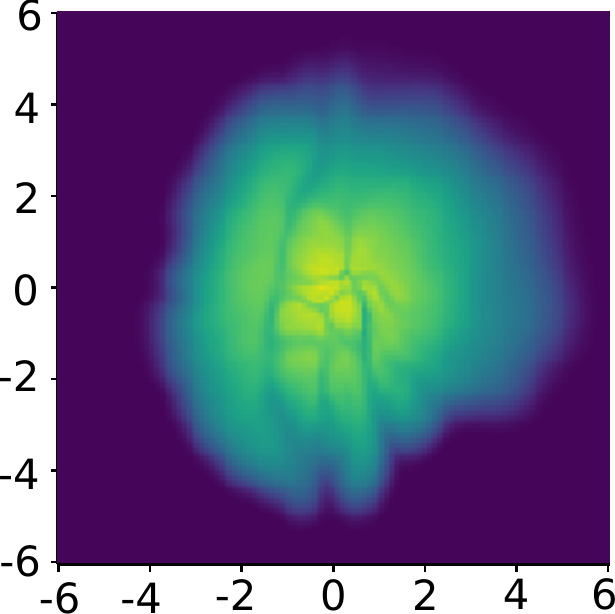}
        & \includegraphics[width=2cm, height=2cm]{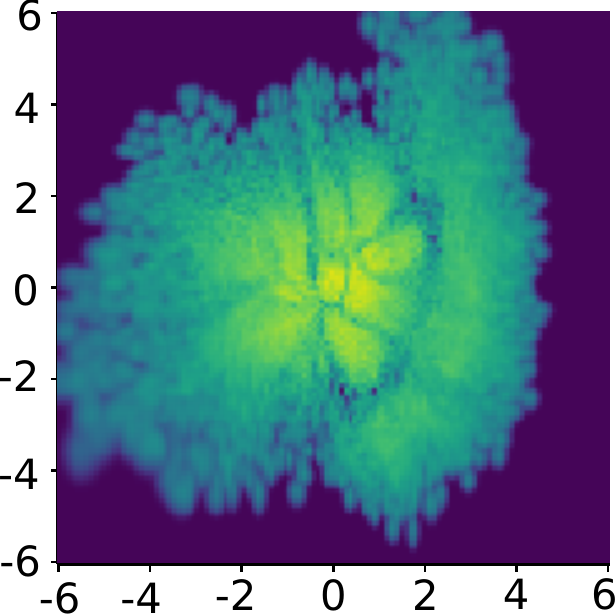} 
        & \includegraphics[width=2cm, height=2cm]{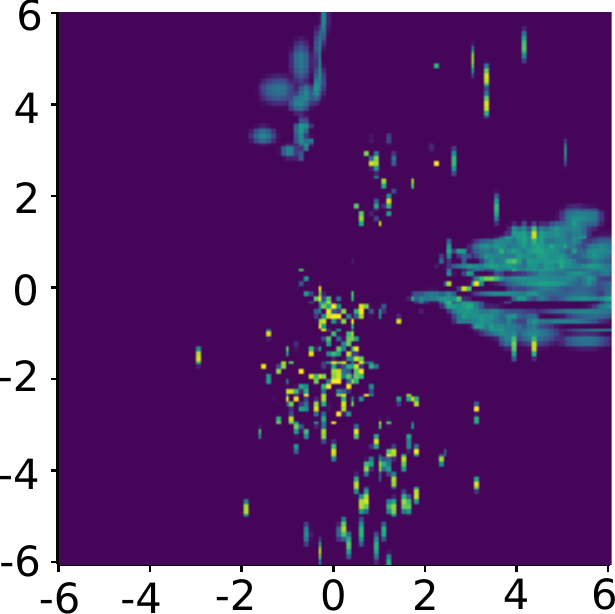}
        & \multirow{2}{2.5cm}{\includegraphics[width=2cm, height=2.6cm]{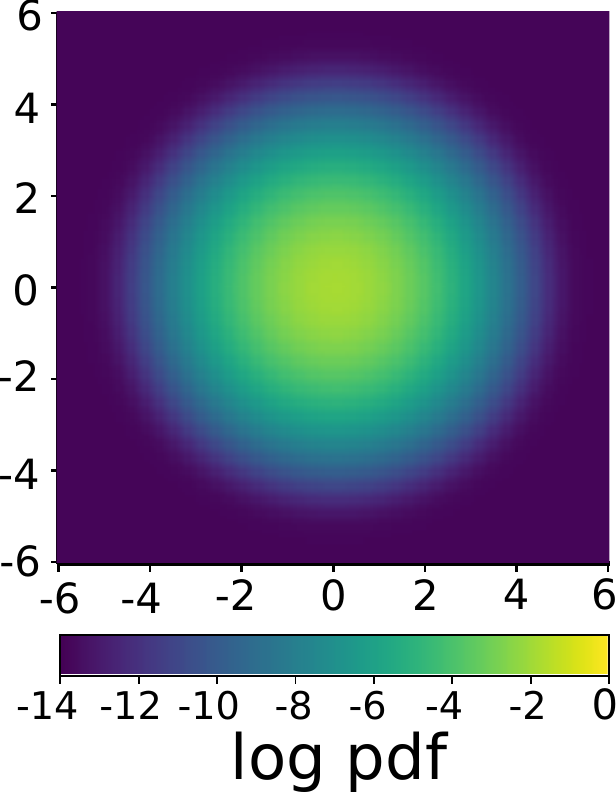}} \\
        %\hline
        Approximate aggregate posterior %$\hat{q}(\rvz)$  
        & \includegraphics[width=2cm, height=2cm]{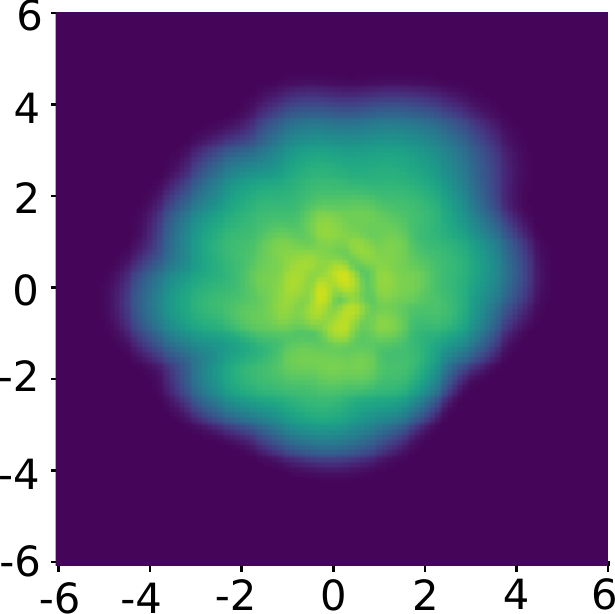}  
        & \includegraphics[width=2cm, height=2cm]{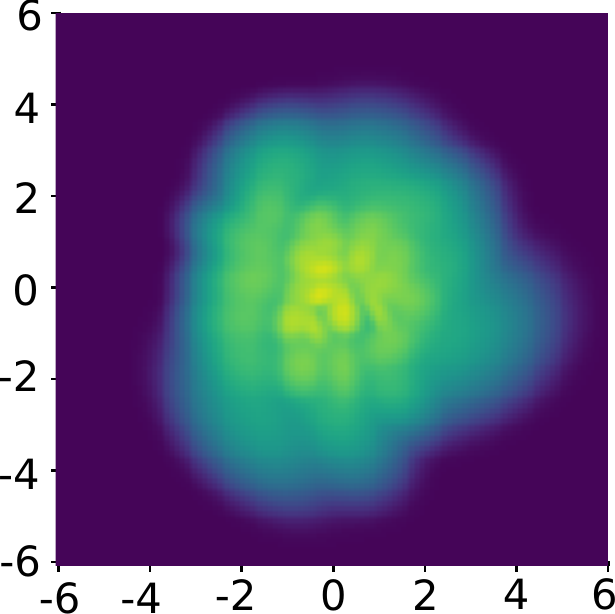}  
        & \includegraphics[width=2cm, height=2cm]{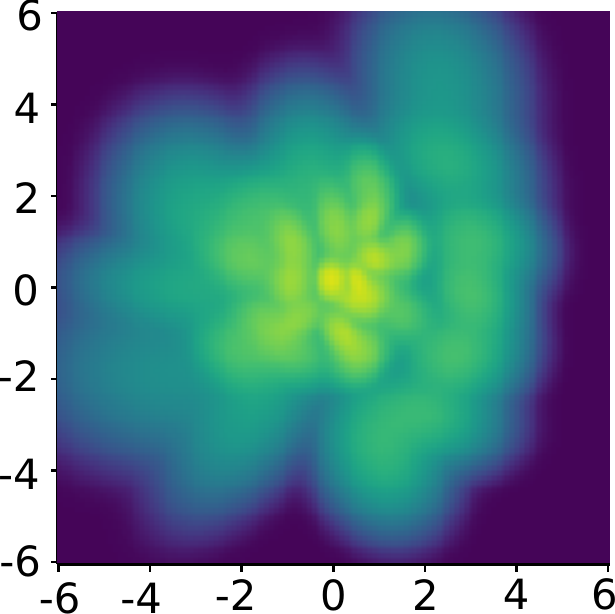} 
        & \includegraphics[width=2cm, height=2cm]{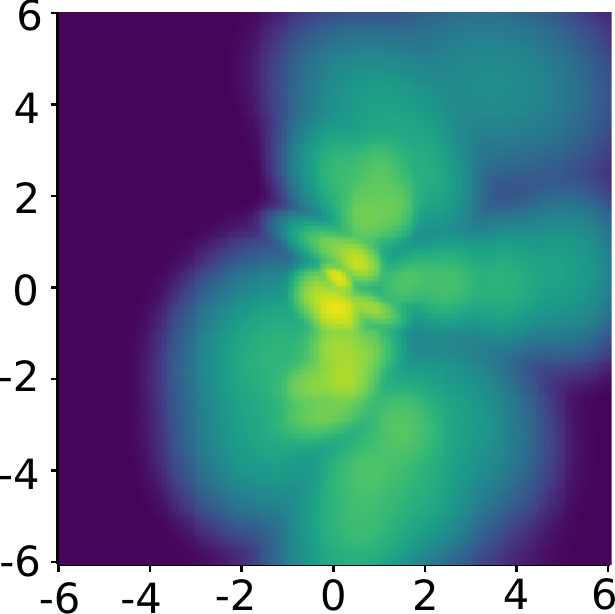} & \\
        \xrowht{10pt}%\cline{1-4}
        $\KL(\hat{q}(\rvz) || p(\rvz))$  & 0.094 $\pm$ 0.003 & 0.132 $\pm$ 0.003
        & 0.422 $\pm$ 0.011 &  1.552 $\pm$ 0.034 &  \\\hline
    \end{tabular}
    \label{tab:MNIST-posterior}
\end{table*}
We carry out extensive experiments on MNIST \cite{MNIST}, Fashion MNIST \cite{fashion-mnist} and CelebA \cite{CelebA-Dataset} datasets. For all datasets, we take the images as real-valued data so that our Gaussian likelihood assumption is appropriate. We take the original VAE and $\beta$-VAE as baseline methods for comparison. We also compare to WAE \cite{WAE} and DIP-VAE \cite{DIP-VAE} to demonstrate the significant performance gain given by optimising the variance parameter and sampling from the approximate aggregate posterior. More results and details of data pre-processing and model architectures are given in Supplemental Materials.  

\subsection{Intuition through visualisation}
First we try to gain an intuitive understanding of the impact of learning the $\sigma^2$ parameter in our proposal. To this end, we trained VAE models of two dimensional latent space on MNIST dataset under different $\sigma^2$ values and in Table \ref{tab:MNIST-posterior} we show log density plots of various distributions in the latent space. 

When $\sigma^2$= 0.5, the weighting $\frac{1}{2\sigma^2}$ between the reconstruction loss and the prior constraint is equal and this corresponds to the original VAE learning objective. When $\sigma^2=1$, the weight on the reconstruction loss is halved, resulting in an increased penalty on the prior regularisation, and this leads to the $\beta$-VAE's objective. $\sigma^2$= 0.035 is learnt under our proposal by optimising the ELBO in (\ref{eq:ELBO-with-sigma}), which learns the optimal balance between the two losses. Finally, $\sigma^2$= 0.01 indicates the scenario where extreme penalty on the reconstruction loss is imposed. 

From the visualisation, we can see that learning under different $\sigma^2$ values leads to very different inference models. 
Larger $\sigma^2$ values result in an aggregate posterior closer to the prior distribution and often corresponding to a smooth density. As the $\sigma^2$ gets smaller, sample posteriors become more distinctive and the aggregate posterior clearly becomes more complex and contains more sophisticated low density regions. Either end of the $\sigma^2$ value spectrum is sub-optimal: too large $\sigma^2$ causes severe information loss of the input data and too small $\sigma^2$ leads to a marginal distribution that is overly complex. We argue that the optimised $\sigma^2$ offers the best balance. More experimental evidence for this claim is given in Section \ref{Sec:performance-gain}.

\begin{figure*}[t]
    \captionsetup[subfigure]{justification=centering}
    \centering
    \begin{subfigure}[t]{0.32\textwidth}
        \centering
        \includegraphics[width=0.95\columnwidth]{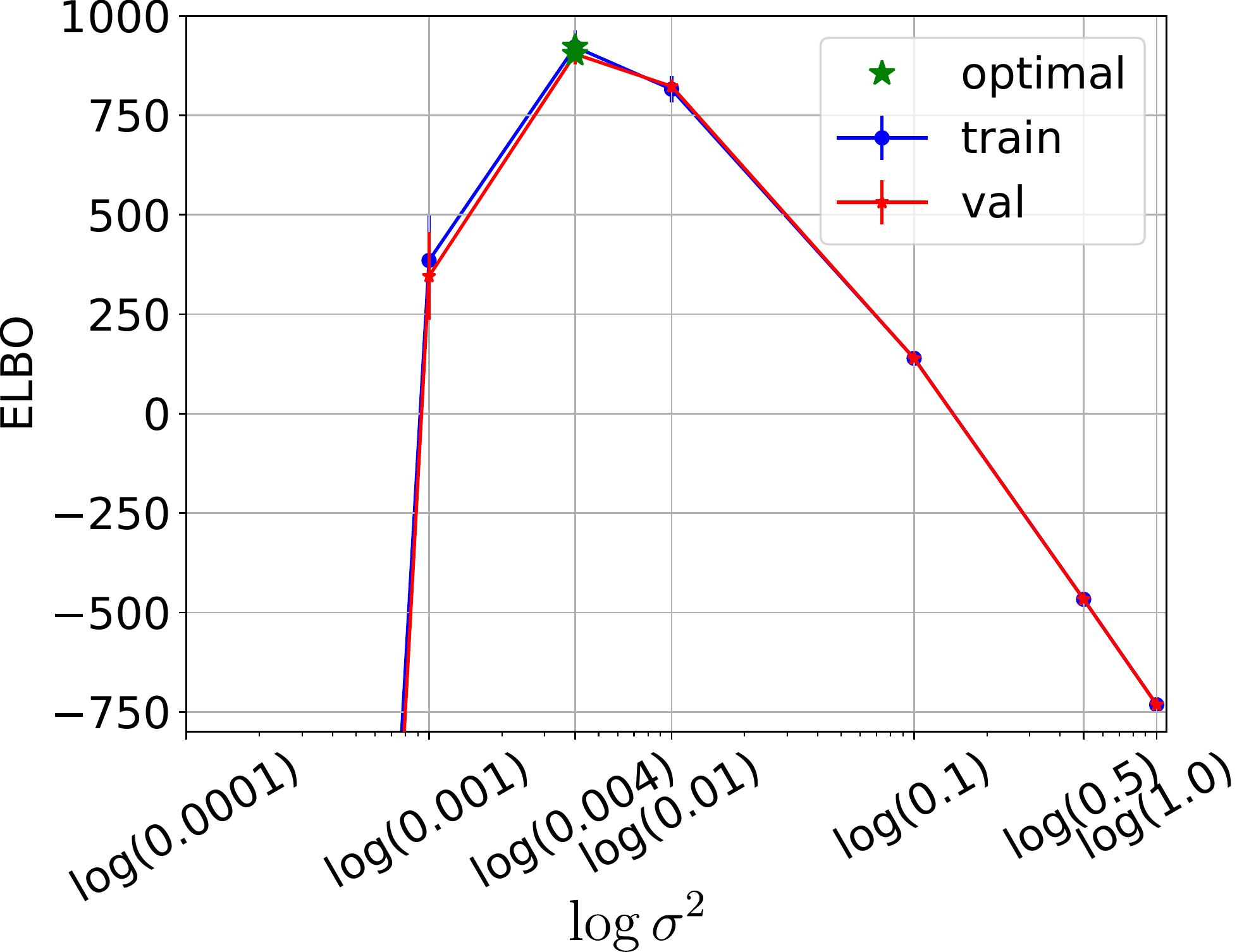}
        \caption{Optimised ELBO (higher is better).}
        \label{fig:ELBO-sigma}
    \end{subfigure}%
    \begin{subfigure}[t]{0.32\textwidth}
        \centering
        \includegraphics[width=0.95\columnwidth]{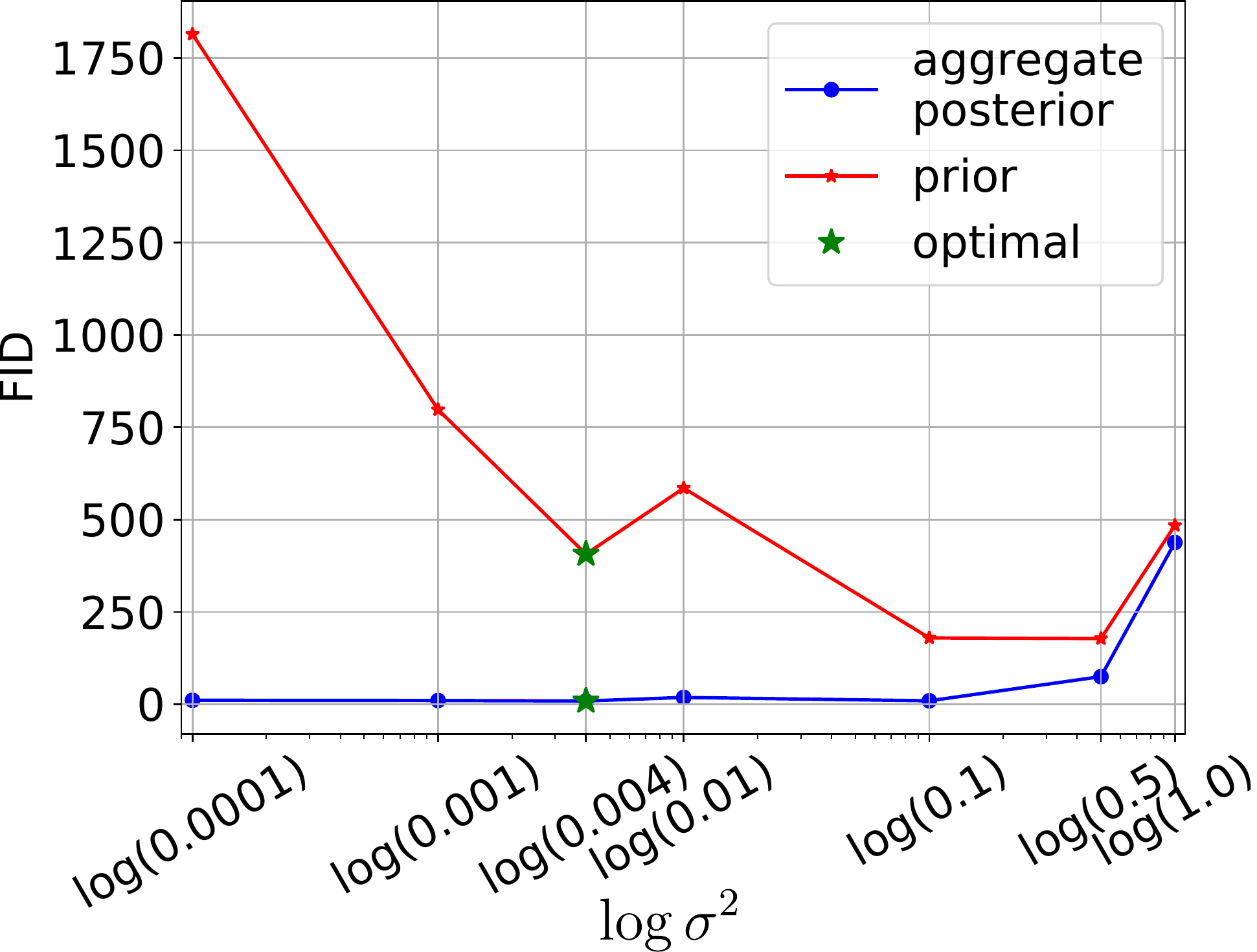}
        \caption{FID score (lower is better).}
        \label{fig:FID-sigma}
    \end{subfigure}%
    \begin{subfigure}[t]{0.32\textwidth}
        \centering
        \includegraphics[width=0.9\columnwidth]{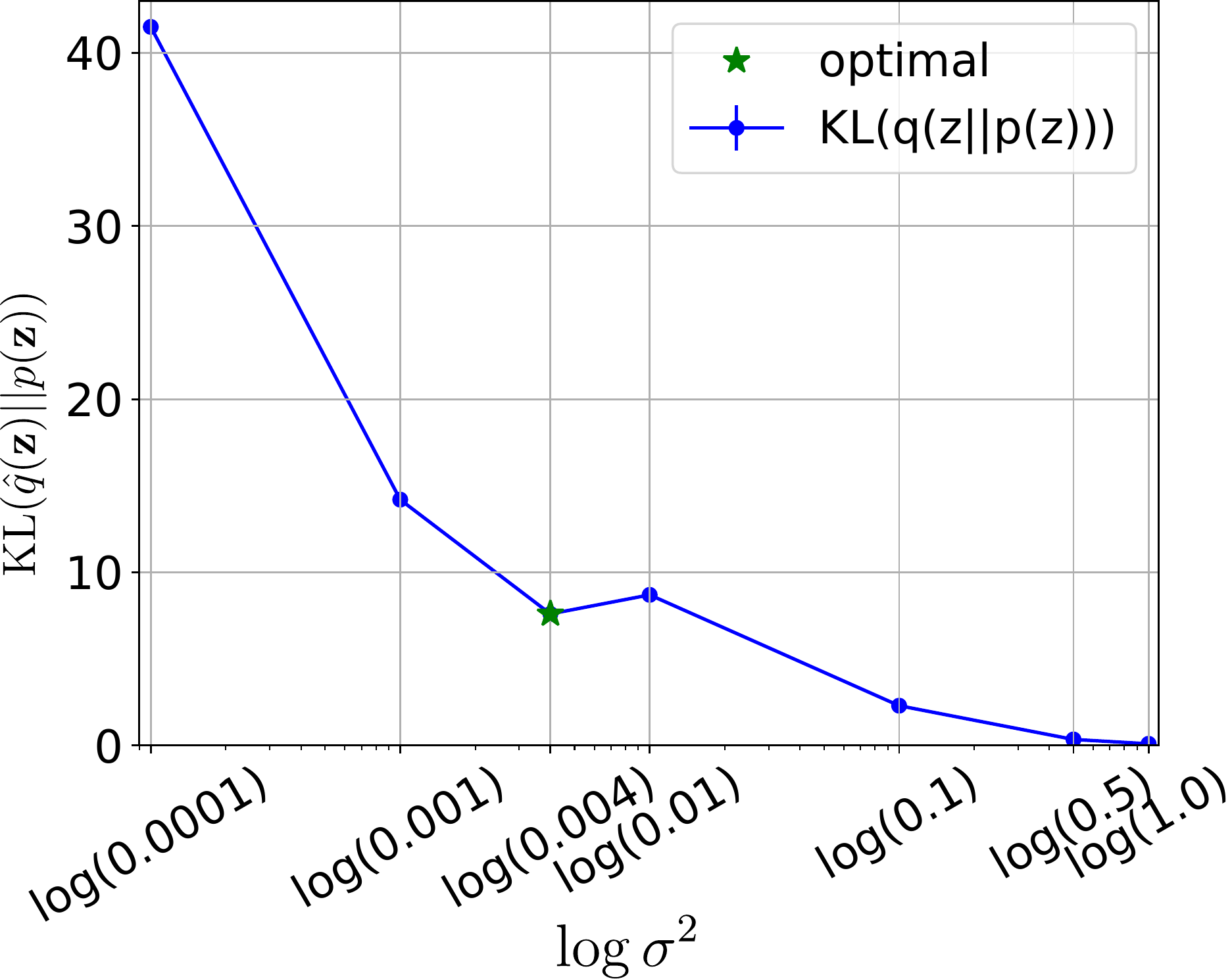}
        \caption{$\KL(\hat{q}(\rvz) || p(\rvz))$ (lower is better).}
        \label{fig:KL-sigma}
    \end{subfigure}
    \caption{Impact of learning $\sigma^2$ on 3 metrics for VAE performance on MNIST dataset. Optimising $\sigma^2$ (green star) reaches the highest ELBO, very low in FID score and reasonably small $\KL(\hat{q}(\rvz) || p(\rvz))$.}
    \label{fig:Qualitative-dataset-sigma}
\end{figure*}

\newcolumntype{C}{ >{\centering\arraybackslash} m{2cm} }
\begin{table*}[t]
    \centering
    \caption{FID scores compared across 5 different methods for MNIST, fashion MNIST and CelebA datasets. For all datasets, optimising $\sigma^2$ achieves the best visual quality (sample are  generated from approx. aggregate posterior where possible).}
    \begin{tabular}{ |B|C|C|C|C|C|C| } 
        \hline\xrowht{10pt}
         & Ours (optimising $\sigma^2$) & VAE \cite{VAE} & $\beta$-VAE \cite{beta-VAE} & DIP-VAE \cite{DIP-VAE} & WAE \cite{WAE} & Real images \\
        \hline
        MNIST  & \textbf{8.9}  & 74.9 & 438.1 & 114.9 & 142.8 & 1.9 \\
        \hline
        Fashion MNIST & \textbf{10.6} & 123.8 & 281.5 & 103.6 & 149.5 & 0.7 \\ \hline 
        CelebA & \textbf{74.7} & 78.4 & 106.4 & 87.5  & 85.4  & 2.9 \\ 
        \hline  
    \end{tabular}
    \label{tab:FID-comparison}
\end{table*}

\newcolumntype{D}{ >{\centering\arraybackslash} m{1.2cm} }
\newcolumntype{E}{ >{\centering\arraybackslash} m{4cm} }
\begin{table*}[t]
    \centering
    \caption{IWAE test log likelihood (LL, higher is better) for MNIST and Fashion MNIST compared across 4 different methods. In our proposal, we evaluate the LL for two different variance estimation settings: 1) a global variance of the entire dataset and 2) input dependent pixel level variances.}
    \begin{tabular}{ B|E|DDDDD } 
        % \hline
        \multicolumn{2}{c}{}& \multicolumn{2}{c}{Optimising $\sigma^2$ (Ours)} & \multirow{2}{1cm}{VAE} & \multirow{2}{1.2cm}{$\beta$-VAE} & \multirow{2}{1.5cm}{DIP-VAE} \\ \cline{3-4}
        \multicolumn{2}{c}{}& dataset & pixel & & & \\ \hline
        \multirow{2}{2.5cm}{MNIST} & Prior               & 1307 & 6037 & -462 & -730 & -461  \\ 
        & Approx. Agg. Posterior & 1323 & \textbf{6052} & -461 & -730 & -460  \\ \hline
        \multirow{2}{2.5cm}{Fashion MNIST} & Prior               & 1145 & 5708 & -463 & -731 & -463  \\ 
        & Approx. Agg. Posterior & 1157 & \textbf{5724} & -462 & -731 & -462  \\ \hline
    \end{tabular}
    \label{tab:IWAE-comparison}
\end{table*}

The visualisation also demonstrates that except for extremely small $\sigma^2$ values, the Gaussian mixture model can effectively approximate the aggregate posterior. Thanks to the closed-form solution of the global variance parameter derived in (\ref{eq:global-sigma-closed-form}) and the stabilised learning procedure introduced in Section \ref{Sec:pixel-sigma}, we will never enter these small $\sigma^2$ values regime in our learning scheme. Furthermore, the gap between the aggregate posterior and the prior persistently exists no matter what value $\sigma^2$ takes. There are good texts explaining why such a gap appears \cite{VAE-suboptimality}. We would like to stress that if such a gap is hard to avoid, then samples should not be naively generated from the prior distribution but from an alternative distribution that better represents the aggregate posterior. We delegate Section \ref{Sec:agg-post-vs-prior} for more detailed discussion on this gap.

\subsection{Performance Gain from Optimizing $\sigma^2$} \label{Sec:performance-gain}
Here we compare the learned VAE models under 3 metrics: 1) the optimised ELBO value (higher indicates a closer bound to the data log likelihood), 2) FID score (lower indicates a better visual quality of the generated samples \cite{FID1}) and 3) the KL divergence between the approximate aggregate posterior and the prior (smaller often indicates a simpler and smoother marginal latent distribution). In Figure \ref{fig:Qualitative-dataset-sigma}, we show these metrics with $\sigma^2$ fixed at 6 different values ranging between 0 and 1 and also trained a VAE model with $\sigma^2$ learnt under the ELBO in (\ref{eq:ELBO-with-sigma}) for the MNIST dataset. When $\sigma^2$ is optimised (indicated by a green star), the model achieves the highest ELBO, one of the lowest FID scores and still remains reasonably close to the prior distribution. In Figure \ref{fig:FID-sigma}, we plot the FID scores evaluated with samples drawn from the approximate posterior (blue) and the prior (red). For all cases the sample quality from aggregate posterior is better than that from the prior distribution and the gap grows exponentially as $\sigma^2$ gets smaller. %This again indicates the importance of generating samples from a distribution that is closer to the aggregate posterior instead of the prior for obtaining good quality samples.   

We also compared the model performance under our proposal (learning $\sigma^2$) with four other learning methods, including the original VAE ($\sigma^2$=0.5), $\beta$-VAE (($\sigma^2$=1), DIP-VAE (\cite{DIP-VAE}) and WAE \cite{WAE} in terms of the generated sample quality measured by FID scores for all three datasets. All methods adopt the same model architecture. The result is given in Table \ref{tab:FID-comparison}. For all datasets, optimising $\sigma^2$ achieves the best FID score. Examples of generated samples are given in Supplemental Materials. 

\begin{figure*}[t]
    \captionsetup[subfigure]{justification=centering}
    \centering
    \begin{subfigure}[t]{0.5\textwidth}
        \centering
        \includegraphics[width=0.95\columnwidth]{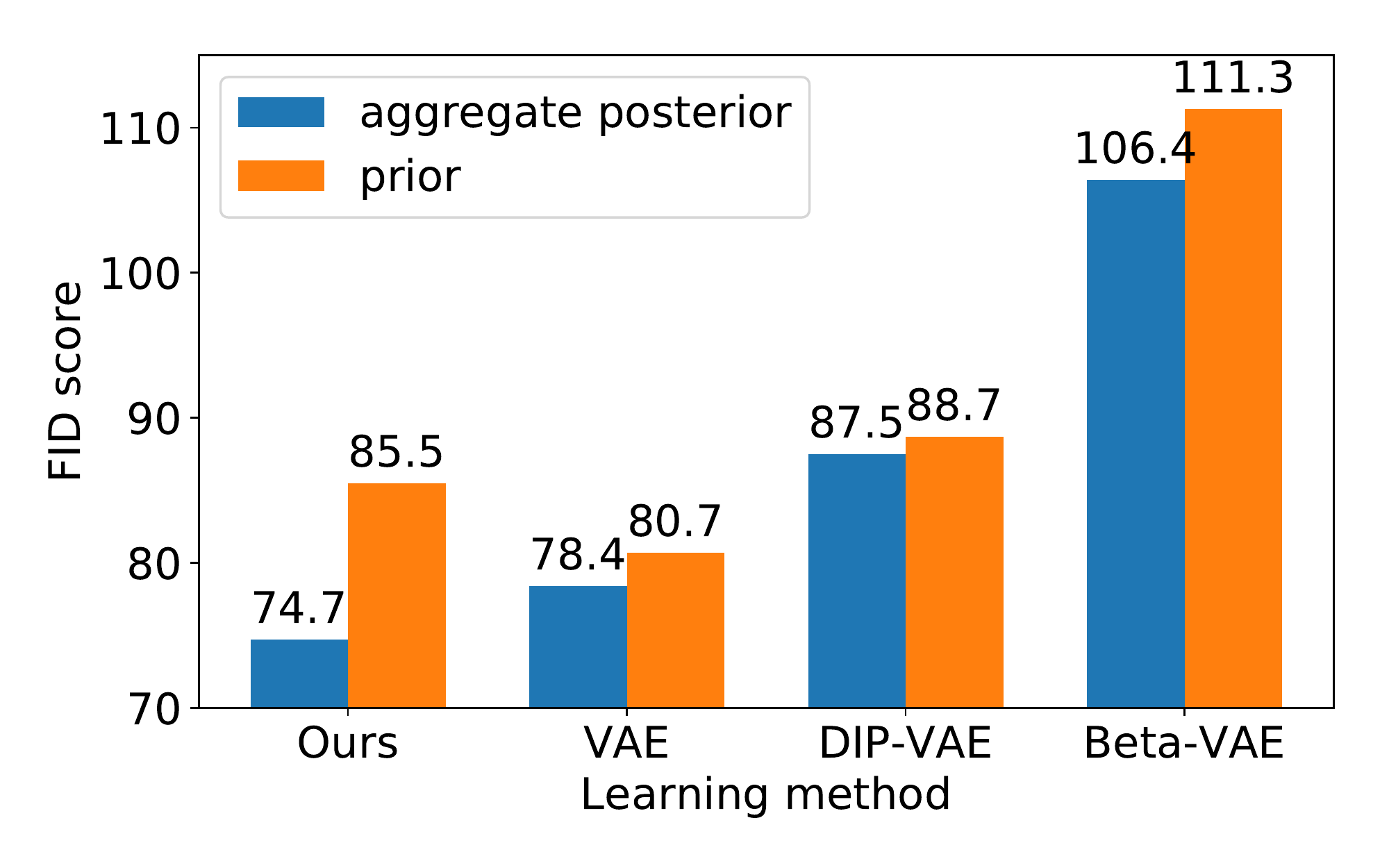}
        \caption{A significant gap between the aggregate posterior (blue) and\\prior (orange) distributions consistently exists across all methods.}
        \label{fig:FID-agg_prior-gap}
    \end{subfigure}%
    \begin{subfigure}[t]{0.5\textwidth}
        \centering
        \includegraphics[width=0.9\columnwidth]{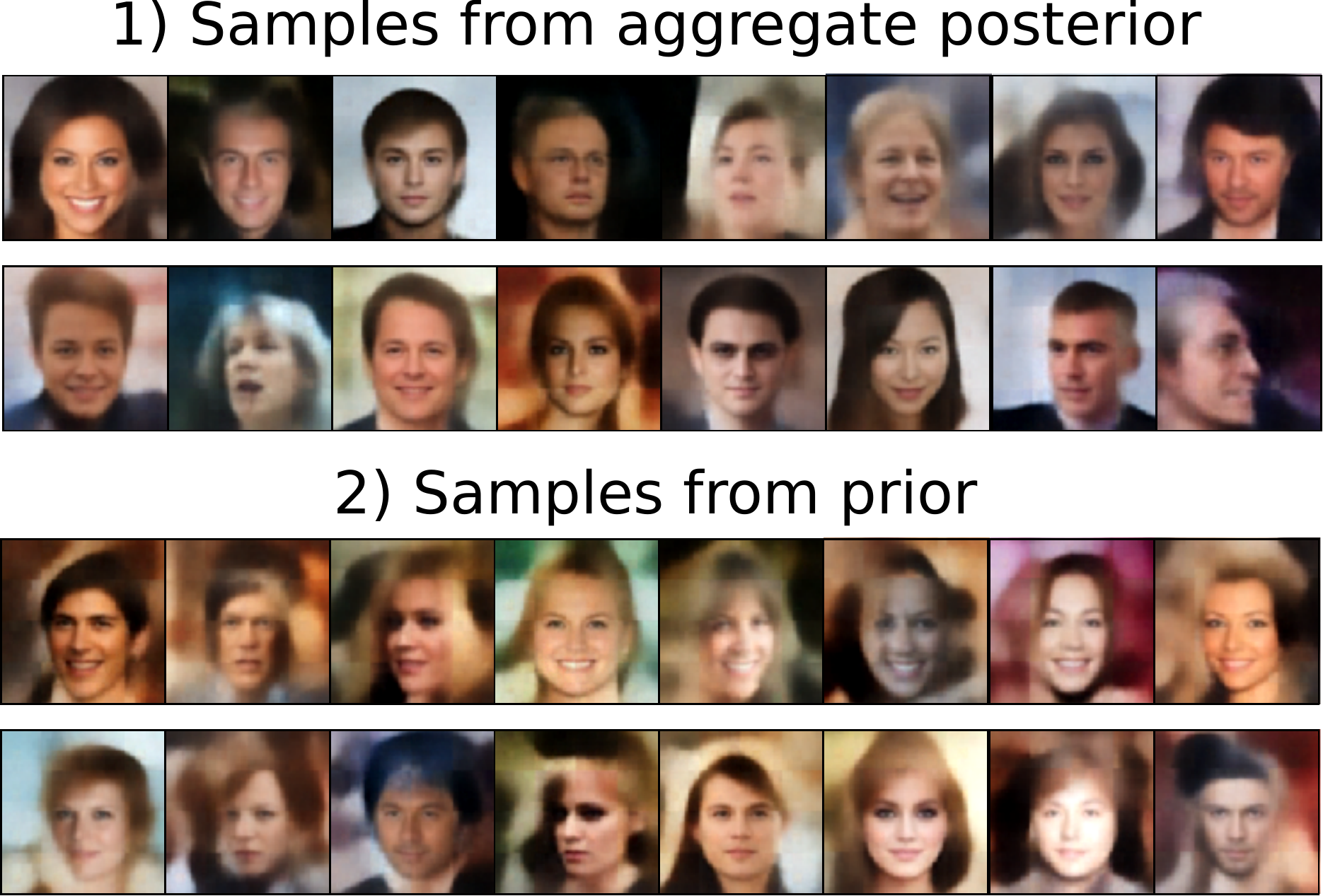}
        \caption{Samples generated from approximate aggregate posterior (top)\\or prior distribution (bottom) for CelebA dataset.}
        \label{fig:celeba-samples-agg-prior-gap}
    \end{subfigure}
    \caption{Gap between aggregate posterior and prior distributions causes bad quality samples to be generated from the prior.}
    \label{fig:gap-agg-prior}
\end{figure*}

Finally, for VAE models with relatively low latent dimensions, we are able to evaluate the IWAE test log likelihood (LL) \cite{IWAE}, which is proposed as a tighter lower bound to the log likelihood $\log p(\rvx)$ than ELBO and hence will give a better evaluation on how well the model is optimised in terms of approaching the data log likelihood objective. In Table \ref{tab:IWAE-comparison}, we list the IWAE test LL evaluated over 500 test data samples with 20 importance points per sample for the MNIST and fashion MNIST datasets where a VAE model of 16-dimensional latent space is learnt. Optimising $\sigma^2$ significantly improves the test LL in comparison to other learning methods. More interestingly, there is a huge performance boost when $\sigma^2$ is estimated across data dimensions and by conditioning on the input data compared to optimising a single-valued $\sigma^2$. Therefore, we conclude that using the full learning procedure to predict a more flexible $\sigma^2$ estimator gives better results.

\subsection{Aggregate Posterior vs Prior} \label{Sec:agg-post-vs-prior}
In Figure \ref{fig:FID-agg_prior-gap}, we compared FID scores of generated samples from both approximate aggregate posterior and prior distributions across 4 different methods and it indicates the sample quality is always better if the samples are drawn from the approximate aggregate posterior. In Figure \ref{fig:celeba-samples-agg-prior-gap}, samples from both distributions are generated under our proposal for the CelebA dataset. The samples from the aggregate posterior are consistently more realistic, whereas the samples from the prior distribution are either with essential facial feature missing or damaged. %Notice a couple of samples from the prior distribution look decent (e.g. first one in top row from the left). These samples are likely generated from a region that the prior and aggregate posterior overlap and unfortunately all the rest samples are generated from the mismatched region between the two distributions. From this observation, it is apparent that we should acknowledge the existence of the gap between the two distributions and to generate good quality samples, we should resort to a distribution that is closer to the aggregate posterior, such as the Gaussian mixture estimate as we suggest in this paper.    

\subsection{Uncertainty Estimation}
\begin{figure}[!ht]
    \captionsetup[subfigure]{justification=centering}
    \centering
    \begin{subfigure}[t]{0.95\columnwidth}
        \centering
        \includegraphics[width=0.98\columnwidth]{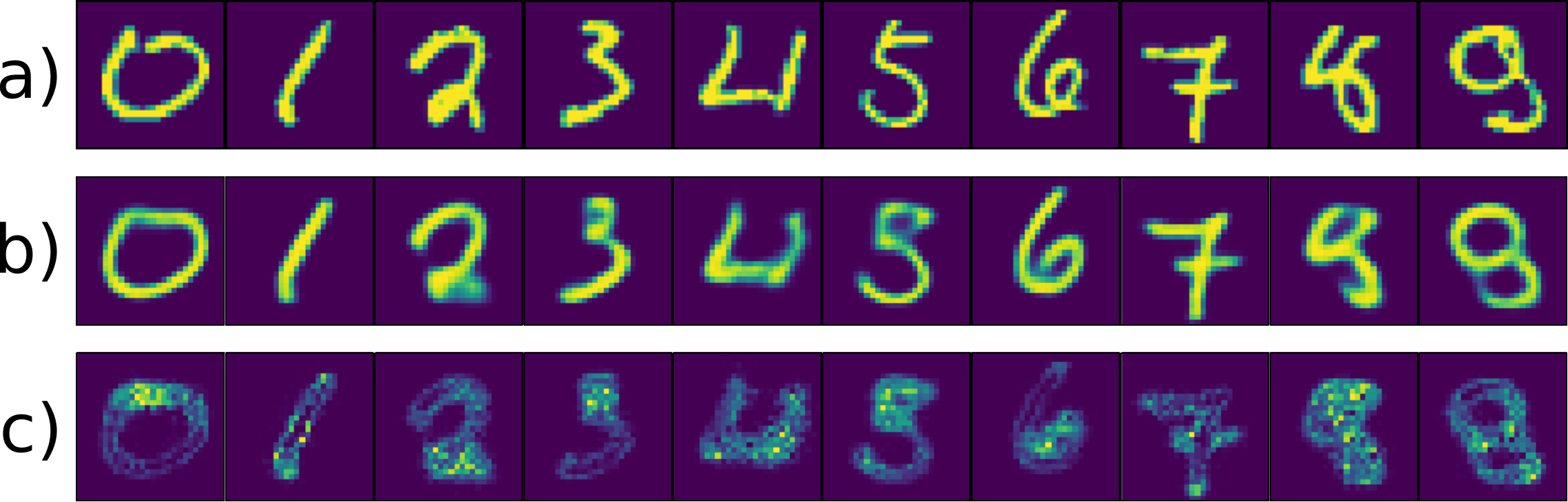}
        \caption{MNIST.}
        \label{fig:MNIST-recons-var}
    \end{subfigure}
    \begin{subfigure}[t]{0.95\columnwidth}
        \centering
        \includegraphics[width=0.98\columnwidth]{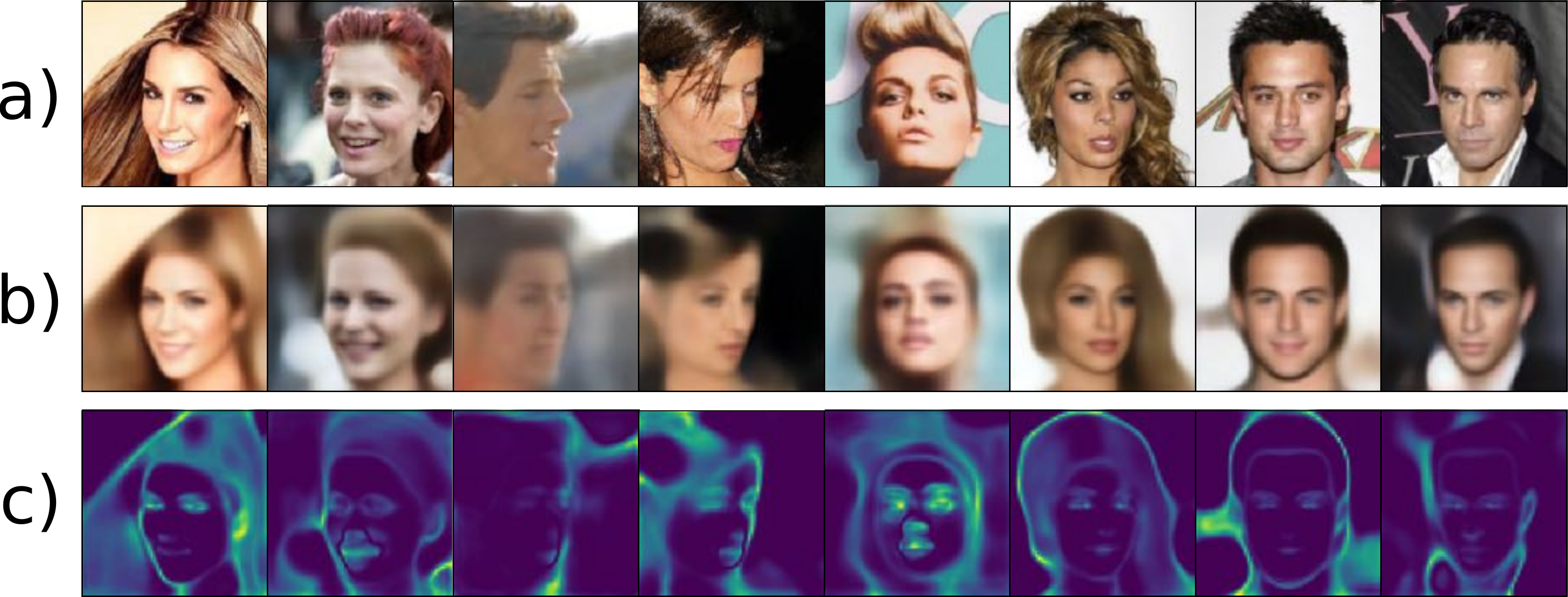}
        \caption{CelebA.}
        \label{fig:celeba-recons-var}
    \end{subfigure}%
    \caption{Uncertainty estimation for reconstructed data. a) ground truth images, b) reconstructed images and c) estimated $\sigma^2$ indicating uncertainty.}
    \label{fig:reconstruction-uncertainty}
\end{figure}

\begin{figure}[!ht]
    \captionsetup[subfigure]{justification=centering}
    \centering
    \begin{subfigure}[t]{0.95\columnwidth}
        \centering
        \includegraphics[width=0.92\columnwidth]{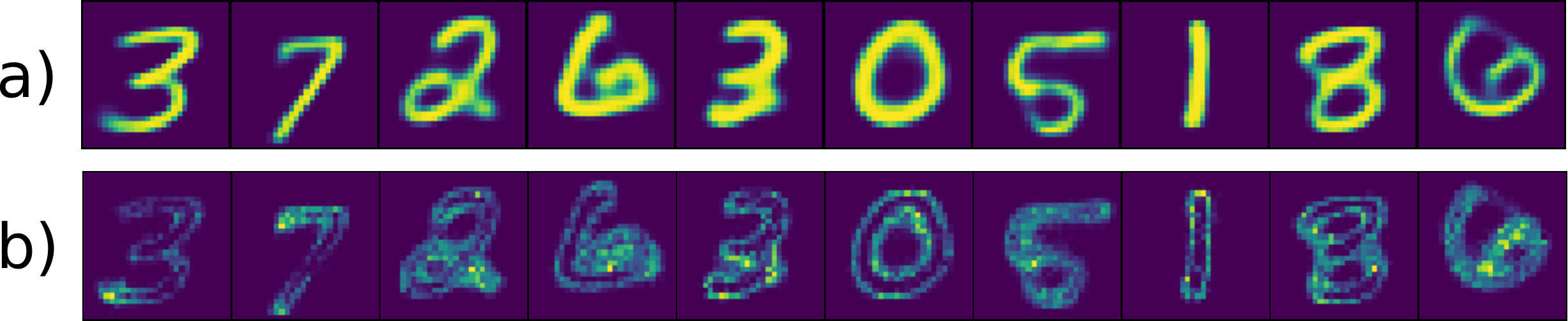}
        \caption{MNIST.}
        \label{fig:MNIST-gen-var}
    \end{subfigure}
    \begin{subfigure}[t]{0.95\columnwidth}
        \centering
        \includegraphics[width=0.95\columnwidth]{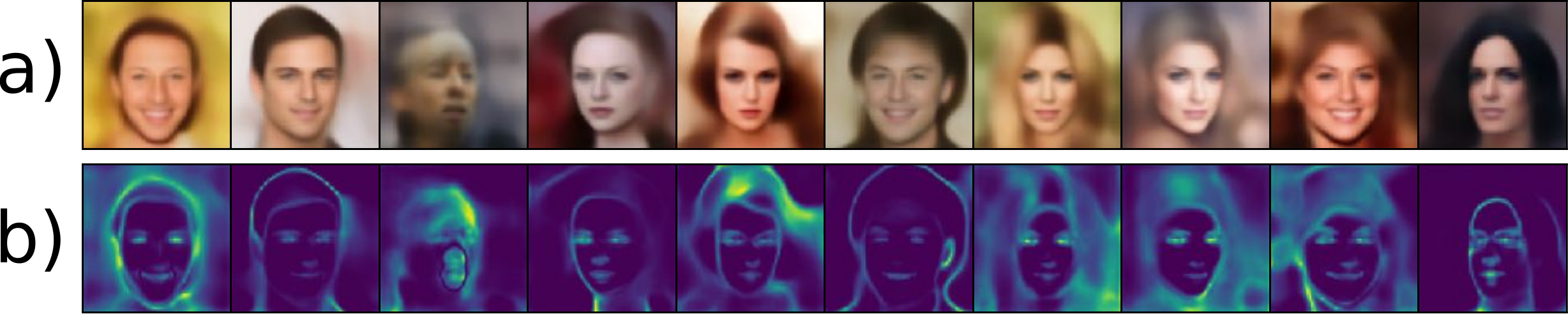}
        \caption{CelebA.}
        \label{fig:celeba-gen-var}
    \end{subfigure}%
    \caption{Uncertainty estimation for generated data. a) generated samples and b) estimated uncertainty.}
    \label{fig:generation-uncertainty}
\end{figure}

A very powerful application of optimising the $\sigma^2$ parameter, especially with the more flexible $\sigma^2$ setting where the $\sigma^2$ is conditioned on a latent encoding and allowed to be variable across data dimensions, is to estimate uncertainty in the generated samples. Figure \ref{fig:reconstruction-uncertainty} illustrates the estimated uncertainty in the reconstructed samples for MNIST and CelebA datasets. Note the predicted $\sigma^2$ highlights the region where the reconstruction is different from the original data samples. During training, $\sigma^2$ is learnt to suppress the difference between the reconstructed and the original data samples and such difference is highly correlated with the statistical irregularity in the dataset. For example, the major difference between the reconstructed digit zero and its corresponding data sample is at the top where the stroke starts and ends. It is the most unpredictable part of writing a zero and that's why the model fails to reconstruct this part accurately. Figure \ref{fig:generation-uncertainty} illustrates uncertainty estimation in generated samples for both MNIST and CelebA datasets. Notice in Figure \ref{fig:celeba-gen-var}, the high uncertainty regions correspond to eyes, mouths, edges of hair and backgrounds, which are highly variable across the dataset. Such uncertainty estimation indicates that the model is aware that there are many other possibilities to generate these regions and it only renders one of them.

\subsection{Ablation Study}
A very important hyperparameter in our learning procedure is the number of components in the Gaussian mixture model that we use to estimate the aggregate posterior $q_{\phi}(\rvz)$. We carried out a study about its impact on the quality of generated samples measured by FID scores for the CelebA dataset and the result is shown in Table \ref{tab:GM-comparison}. Increasing the number of components does not lead to any obvious improvement in the generation quality (FID score is only improves by 1.5 ($ < 2\%$) when the number of mixtures increases from 500 to 2000). Only when the number of mixture is extremely large, such as 10k, we can see a significant gain, but this is likely to be caused by the approximation becoming overfitted to the 50k samples that we use to fit the mixture model. Therefore, in all of our experiments, we stick with a relatively small number of Gaussian mixtures, specifically 32 for MNIST and fashion MNIST and 500 for CelebA.  
\newcolumntype{G}{ >{\centering\arraybackslash} m{1.7cm} }
\newcolumntype{F}{ >{\centering\arraybackslash} m{1.1cm} }
\begin{table}[H]
    \centering
    \caption{Study the impact of the number of Gaussian mixtures used in the approximation of the aggregate posterior on the generated sample quality (CelebA dataset).}
    \begin{tabular}{ |G|F|F|F|F| } 
        \hline
        N mixtures & 500   & 1000 & 2000 & 10000 \\\hline
        FID          & 79.9  & 79.2 & 78.4 & 73.1 \\\hline  
    \end{tabular}
    \label{tab:GM-comparison}
\end{table}

\section{Conclusion}
In this paper, we propose a learning algorithm that automatically achieves (for real-valued data) optimal balance between the two competing objectives (reconstruction loss and prior regularisation) for the VAE ELBO loss. A convenient by-product of our learning scheme is an effective uncertainty estimator for the generation or reconstruction prediction, allowing a wider range of potential applications, including safety critical environments.
We further study the gap between the aggregate posterior and the prior distribution, which is associated with poor samples being generated, and offer a simple solution to mitigate such problems.

%\section{ Acknowledgments}

\makeatletter
\setlength{\itemindent}{0pt}
\setlength{\labelsep}{0pt}
\setlength{\labelwidth}{0pt}
\let\@biblabel\@gobble
\makeatother

\bibliographystyle{aaai}



\section*{Supplementary material}

# Supplemental Materials

In this appendix, we give more details about the experiments we carried out in this paper, including data pre-processing procedures and model architectures. We also present additional results, which cannot be included in the main content due to the space limit.

## A. Datasets and Pre-processing

We ran experiments on MNIST, Fashion MNIST and CelebA datasets. For all datasets, we take the images as real-valued data.

### A.1. MNIST and Fashion MNIST

MNIST and Fashion MNIST datasets both consist of a training and a test set with 60k and 10k images of hand-written digits, where each image is a $28 \times 28$ grey-scale image. We rescale the original images so that the pixel values lying within the range $[0, 1]$.

### A.2. CelebA

CelebA dataset consists of 202599 RGB images of celebrity faces. We first resize all the images to $128 \times 128$ and rescale the pixel values to be within $[0, 1]$. We then randomly select 20k images (about 10%) to be the test set.

## B. Model Architectures

### B.1. MNIST and Fashion MNIST

We use CNN architectures for both MNIST and Fashion MNIST encoder and decoder network (with additional depth to space operations for the decoder), as detailed below. We use the exactly same network for both datasets.

Encoder network:

$$\boldsymbol{x} \in \mathcal{R}^{28 \times 28 \times 1} \to \text{Symmetric padding} \to \mathcal{R}^{32 \times 32 \times 1} \to \text{Conv}_{16}^{4 \times 4} \ (\text{stride} = 2) \to \text{ReLU}$$
$$\to \text{Conv}_{64}^{4 \times 4} \ (\text{stride} = 2) \to \text{ReLU}$$
$$\to \text{Conv}_{256}^{4 \times 4} \ (\text{stride} = 2) \to \text{ReLU}$$
$$\to \text{FC}_{64} \to \text{FC}_{d_z} \Rightarrow \mu \in \mathcal{R}^{d_z}$$
$$\searrow$$
$$\text{FC}_{d_z} \to \text{ReLU} \Rightarrow \sigma \in \mathcal{R}^{d_z}$$

Decoder network:

$$\boldsymbol{z} = \mu + \epsilon \odot \sigma \to \text{FC}_{4096} \to \text{ReLU} \to \text{Reshape} \to \mathcal{R}^{1 \times 1 \times 4096}$$
$$\to \text{depth to space}_4 \to \mathcal{R}^{4 \times 4 \times 256} \to \text{Conv}_{256}^{3 \times 3} \ (\text{stride} = 1) \to \text{ReLU}$$
$$\to \text{depth to space}_2 \to \mathcal{R}^{8 \times 8 \times 64} \to \text{Conv}_{64}^{3 \times 3} \ (\text{stride} = 1) \to \text{ReLU}$$
$$\to \text{depth to space}_2 \to \mathcal{R}^{16 \times 16 \times 16} \to \text{Conv}_{16}^{3 \times 3} \ (\text{stride} = 1) \to \text{ReLU}$$
$$\to \text{depth to space}_2 \to \mathcal{R}^{32 \times 32 \times 4} \to \text{Conv}_{1}^{5 \times 5} \ (\text{stride} = 1, \text{padding=valid})$$
$$\to \text{Sigmoid} \to \text{Reshape} \Rightarrow \hat{\boldsymbol{x}} \in \mathcal{R}^{28 \times 28 \times 1}$$

Pixel-wise variance estimator network:

$$\mathcal{R}^{8\times8\times64}(\text{from decoder}) \rightarrow \text{Conv}_{128}^{3\times3} \text{ (stride}=1) \rightarrow \text{ReLU}$$
$$\rightarrow \text{depth to space}_2 \rightarrow \mathcal{R}^{16\times16\times32} \rightarrow \text{Conv}_{32}^{3\times3} \text{ (stride}=1) \rightarrow \text{ReLU}$$
$$\rightarrow \text{depth to space}_2 \rightarrow \mathcal{R}^{32\times32\times8} \rightarrow \text{Conv}_1^{5\times5} \text{ (stride}=1, \text{padding=valid)}$$
$$\rightarrow \text{Sigmoid} \rightarrow \text{Reshape} \Rightarrow \hat{x} \in \mathcal{R}^{28\times28\times1}$$

$d_z$ is the code dimension, $\mu$ is code mean and $\sigma$ is code standard deviation, $\epsilon \sim \mathcal{N}(0, \mathcal{I})$ and unless otherwise specified the padding in the convolution operation is 'same'. We use the following hyper-parameters to train the network:

| Batch size | Code size | Number of epochs | | Optimizer | Learning rate | Padding |
|---|---|---|---|---|---|---|
| | | Stage 1 | Stage 2 | | | |
| 256 | 16 | 16 | 14 | Adam | $5 \times 10^{-4}$ | SAME |

## B.2. CelebA

For CelebA dataset, we implement one architecture based on CNN and transposed CNN, as detailed below: Encoder network:

$$x \in \mathcal{R}^{128\times128\times3} \rightarrow \text{Conv}_{32}^{4\times4} \text{ (stride}=2) \rightarrow \text{Leaky ReLU}$$
$$\rightarrow \text{Conv}_{64}^{4\times4} \text{ (stride}=2) \rightarrow \text{Leaky ReLU}$$
$$\rightarrow \text{Conv}_{128}^{4\times4} \text{ (stride}=2) \rightarrow \text{Leaky ReLU}$$
$$\rightarrow \text{Conv}_{256}^{4\times4} \text{ (stride}=2) \rightarrow \text{Leaky ReLU}$$
$$\rightarrow \text{Conv}_{512}^{4\times4} \text{ (stride}=2) \rightarrow \text{Leaky ReLU}$$
$$\rightarrow \text{Conv}_{1024}^{4\times4} \text{ (stride}=2, \text{padding=valid)} \rightarrow \text{Leaky ReLU}$$
$$\rightarrow \text{FC}_{512} \rightarrow \text{FC}_{d_z} \Rightarrow \mu \in \mathcal{R}^{d_z}$$
$$\searrow$$
$$\text{FC}_{d_z} \Rightarrow \sigma \in \mathcal{R}^{d_z}$$

Decoder network:

$$z = \mu + \epsilon \odot \sigma \rightarrow \text{FC}_{256} \rightarrow \text{Leaky ReLU} \rightarrow \text{Reshape} \rightarrow \mathcal{R}^{1\times1\times256} \rightarrow \text{Conv}_{4096}^{1\times1}$$
$$\rightarrow \text{TransposedConv}_{1024}^{4\times4} \text{ (stride}=(1,1)) \rightarrow \text{BN} \rightarrow \text{Leaky ReLU} \rightarrow \mathcal{R}^{1\times1\times1024}$$
$$\rightarrow \text{TransposedConv}_{512}^{4\times4} \text{ (stride}=(4,4)) \rightarrow \text{BN} \rightarrow \text{Leaky ReLU} \rightarrow \mathcal{R}^{4\times4\times512}$$
$$\rightarrow \text{TransposedConv}_{256}^{4\times4} \text{ (stride}=(4,4)) \rightarrow \text{BN} \rightarrow \text{Leaky ReLU} \rightarrow \mathcal{R}^{16\times16\times256}$$
$$\rightarrow \text{TransposedConv}_{128}^{4\times4} \text{ (stride}=(4,4)) \rightarrow \text{BN} \rightarrow \text{Leaky ReLU} \rightarrow \mathcal{R}^{64\times64\times128}$$
$$\rightarrow \text{TransposedConv}_3^{4\times4} \text{ (stride}=(2,2)) \Rightarrow \hat{x} \in \mathcal{R}^{64\times64\times3}$$

Pixel-wise variance estimator network:

$$\mathcal{R}^{4\times4\times512}(\text{from decoder}) \rightarrow \text{TransposedConv}_{256}^{4\times4} \text{ (stride}=(4,4)) \rightarrow \text{Leaky ReLU} \rightarrow \mathcal{R}^{16\times16\times256}$$
$$\rightarrow \text{TransposedConv}_{128}^{4\times4} \text{ (stride}=(4,4)) \rightarrow \text{Leaky ReLU} \rightarrow \mathcal{R}^{64\times64\times128}$$
$$\rightarrow \text{TransposedConv}_1^{4\times4} \text{ (stride}=(2,2)) \Rightarrow \hat{x} \in \mathcal{R}^{128\times128\times1}$$

BN stands for batch norm. We use the following hyper-parameters to train the network:

| Batch size | Code size | Number of epochs | | Optimizer | Learning rate | Padding |
|---|---|---|---|---|---|---|
| | | Stage 1 | Stage 2 | | | |
| 256 | 256 | 75 | 20 | Adam | $5 \times 10^{-4}$ | SAME |

## C. More Visualisation of MNIST 2D Aggregate Posteriors

Visualisation of the marginal latent distribution for MNIST dataset with a 2D latent space is very helpful to understand the impact of different variance values on the learnt marginal distributions. We showed plots for 4 different variance values in the main texts. Here we list more variance values to illustrate more clearly the transition of the learnt marginal posterior (or aggregate posterior - the data distribution in the latent space) from close to prior distribution to quite different when $\sigma^2$ decreases from 1 to very small value. The optimal $\sigma^2$ (0.035, found by maximising ELBO) gives an auto-encoding procedure that maintains the best balance between reconstruction error and prior regularisation.

*Table 1.* Comparison of the aggregate posterior and its Gaussian mixture approximation (32 mixtures) for models learnt under different $\sigma^2$ values for MNIST dataset. $D_{\mathrm{KL}}(\hat{q}(\mathbf{z})\|p(\mathbf{z}))$ denotes the gap between the approximate aggregate posterior and the prior distribution and is evaluated by Monte Carlo estimation of 10k samples (10 runs).

| $\sigma^2$ | 1 | 0.5 | 0.1 | Prior |
|---|---|---|---|---|
| Aggregate posterior | 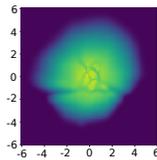 | 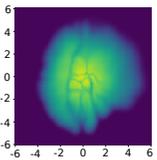 | 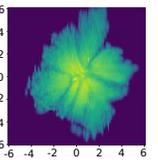 | 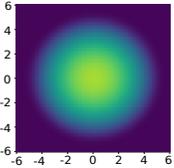 |
| Approximate aggregate posterior | 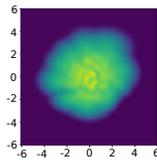 | 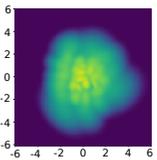 | 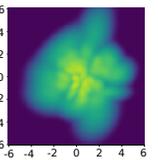 | |
| $D_{\mathrm{KL}}(q(\mathbf{z})\|p(\mathbf{z}))$ | $0.094 \pm 0.003$ | $0.132 \pm 0.003$ | $0.328 \pm 0.009$ | |

| $\sigma^2$ | 0.05 | 0.02 | 0.01 | **0.035 (Optimal)** |
|---|---|---|---|---|
| Aggregate posterior | 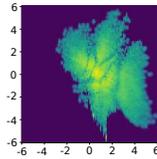 | 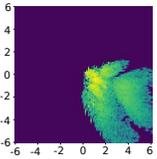 | 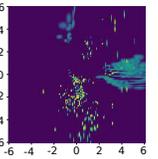 | 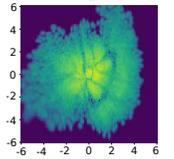 |
| Approximate aggregate posterior | 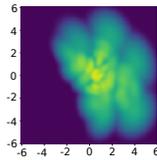 | 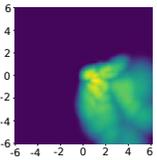 | 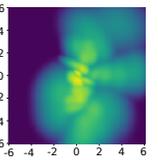 | 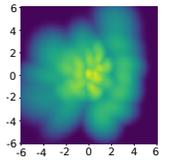 |
| $D_{\mathrm{KL}}(q(\mathbf{z})\|p(\mathbf{z}))$ | $0.812 \pm 0.013$ | $2.143 \pm 0.030$ | $1.552 \pm 0.034$ | $0.422 \pm 0.011$ |

## D. Generated Samples

In the main text, we present some generated samples from our proposal where the variance parameter $\sigma^2$ in the likelihood model is being optimised under the ELBO loss and FID scores of the generated samples given by different methods. Here (Figure 2-13) we show more examples of the generated samples from all 5 different methods, including ours, VAE, $\beta$-VAE, DIP-VAE and WAE for both MNIST and CelebA datasets. For each method, we generate samples from both the approximate aggregate distribution (using Gaussian mixture with 500 components) and the prior distribution.

Judging the quality of generated samples visually is hard, as it can be subjective. Therefore, we think FID scores probably offers a more convincing evaluation and provide that in the main texts. Nonetheless, we think inspecting the generated samples can offer richer information than a single score. Hence, we present random drawn samples here. By comparison, we think the generated samples from our proposal are significantly better than other methods for both datasets, especially in the aspects of diversity and the details that real images present. Across all methods, the generated samples from aggregate posteriors are at better visual quality compared to those from the prior distribution.

The sample posteriors in WAE are extremely sharp Gaussian distributions. Therefore, estimating the aggregate posterior for this method using Guassian mixture model is not sensible. So we only have samples generated from the prior distribution for both MNIST and CelebA datasets for the WAE method, as shown in Figure 1.

### D.1. MNIST

Figure 2-5 illustrates 100 generated samples for MNIST dataset given by our method, VAE, $\beta$-VAE and DIP-VAE respectively.

### D.2. CelebA

Figure 10-13 illustrates 100 generated samples for MNIST dataset given by our method, VAE, $\beta$-VAE and DIP-VAE respectively.

## E. Uncertainty Estimation

We can conveniently use the learnt variance prediction to estimate the uncertainty in the generated or reconstructed samples. We showed some examples in the main text and here we give more results for both MNIST and CelebA datasets in Figure 14-19.

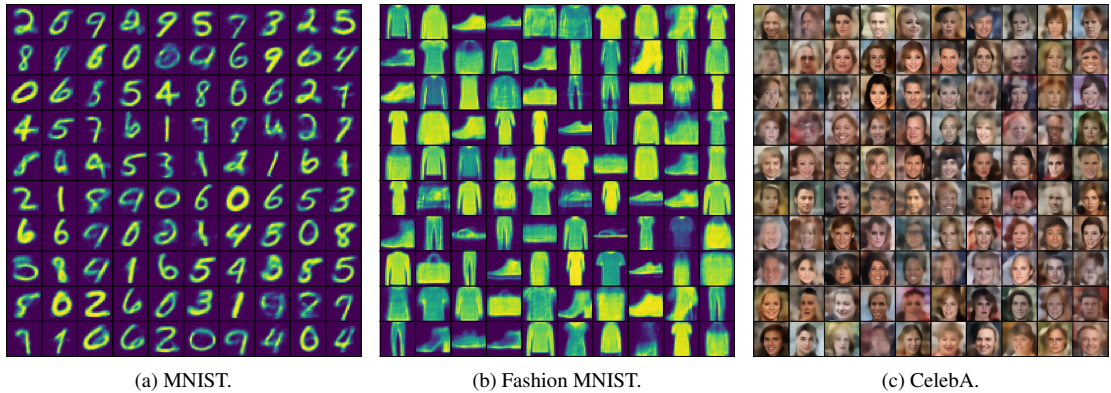

(a) MNIST.  (b) Fashion MNIST.  (c) CelebA.

*Figure 1.* 100 images generated by WAE where the latent distribution of the generated samples is required to match the prior distribution via moment matching or adversarial loss. Samples are generated from prior only, as the aggregate posterior cannot be estimated simply in this method.

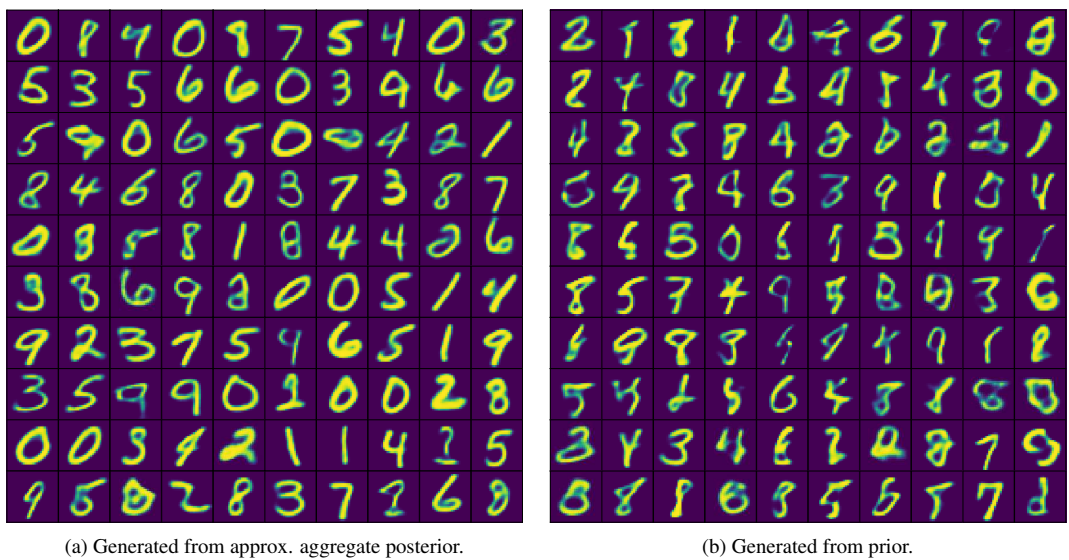

(a) Generated from approx. aggregate posterior.

(b) Generated from prior.

*Figure 2.* 100 MNIST images generated by our proposal where the variance in the likelihood model is learnt.

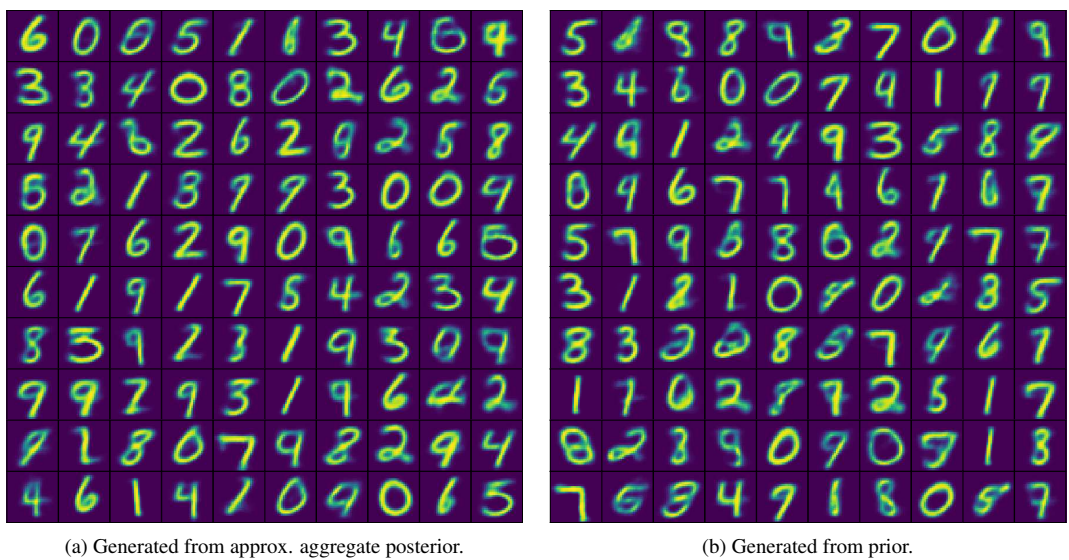

(a) Generated from approx. aggregate posterior.

(b) Generated from prior.

*Figure 3.* 100 MNIST images generated by VAE where the variance in the likelihood model is fixed at 0.5.

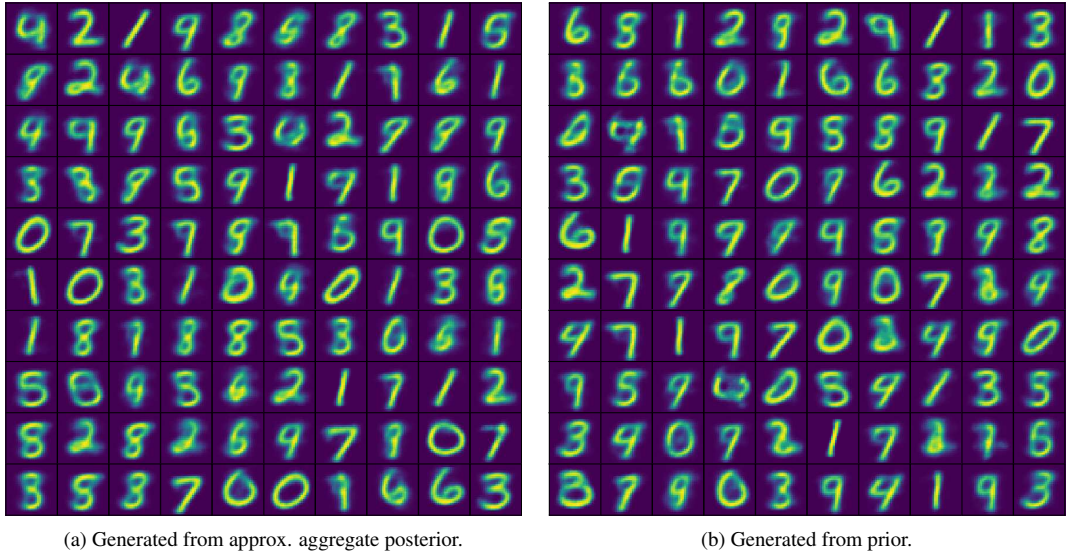

(a) Generated from approx. aggregate posterior.

(b) Generated from prior.

*Figure 4.* 100 MNIST images generated by $\beta$-VAE where the variance in the likelihood model is fixed at 1.

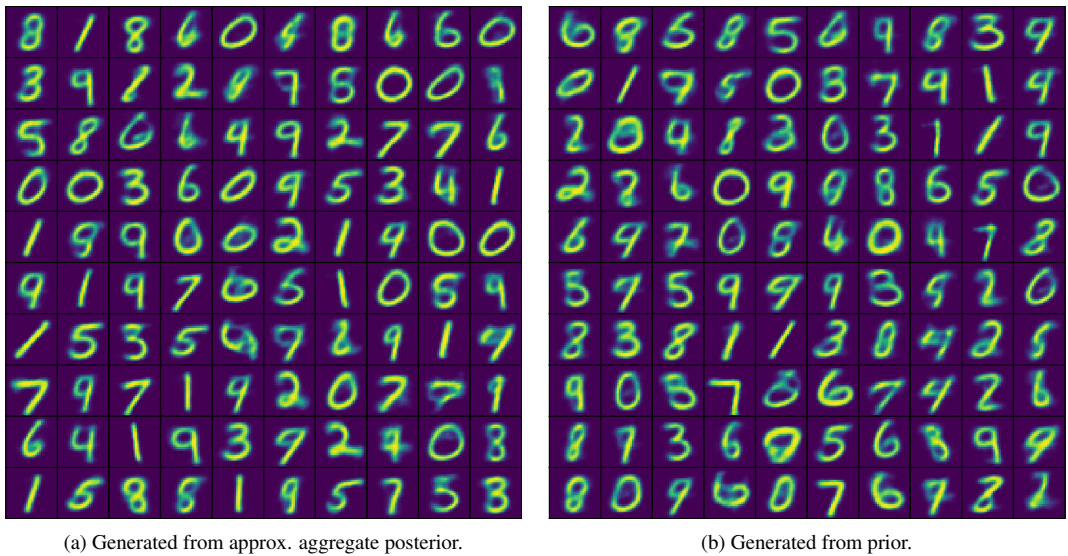

(a) Generated from approx. aggregate posterior.

(b) Generated from prior.

*Figure 5.* 100 MNIST images generated by DIP-VAE where a lower bound of the ELBO loss is being optimised.

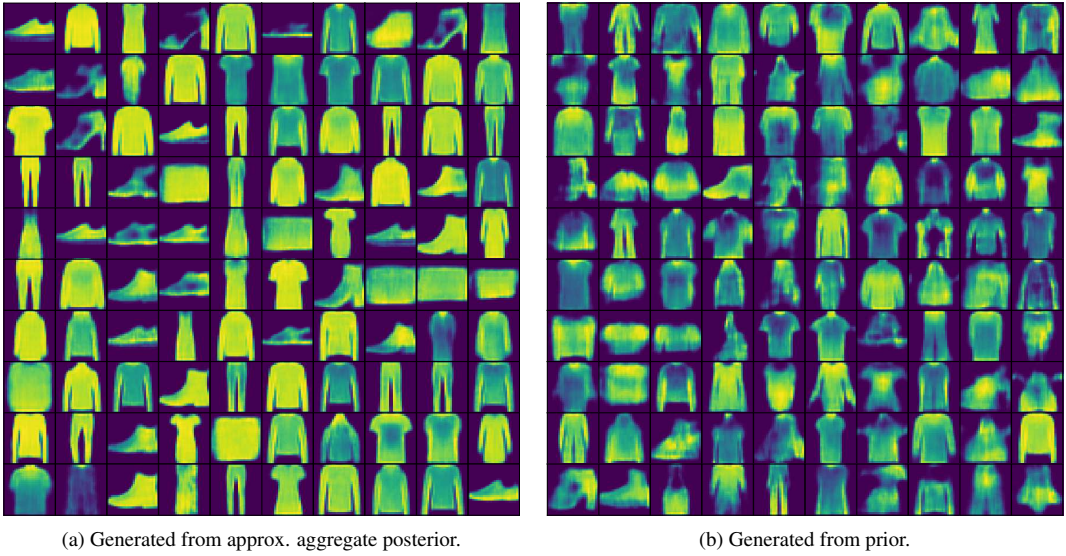

(a) Generated from approx. aggregate posterior.

(b) Generated from prior.

*Figure 6.* 100 Fashion MNIST images generated by our proposal where the variance in the likelihood model is learnt.

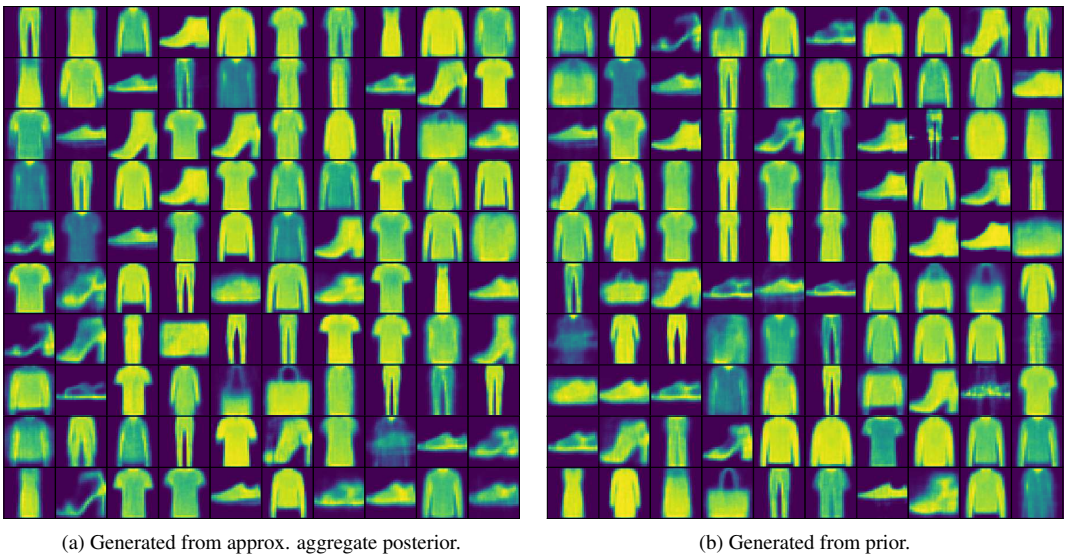

(a) Generated from approx. aggregate posterior.

(b) Generated from prior.

*Figure 7.* 100 Fashion MNIST images generated by VAE where the variance in the likelihood model is fixed at 0.5.

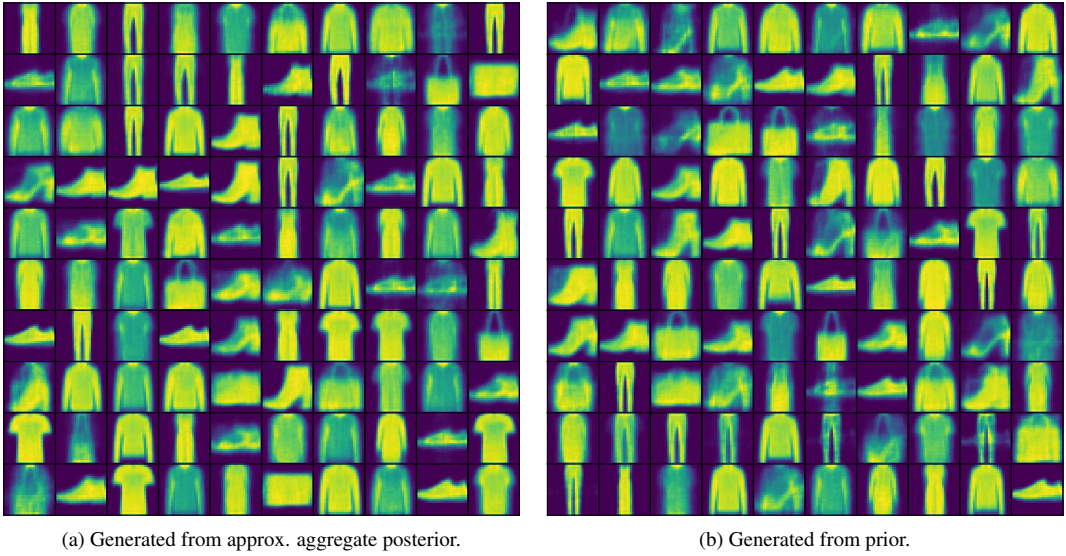

(a) Generated from approx. aggregate posterior.      (b) Generated from prior.

*Figure 8.* 100 Fashion MNIST images generated by $\beta$-VAE where the variance in the likelihood model is fixed at 1.

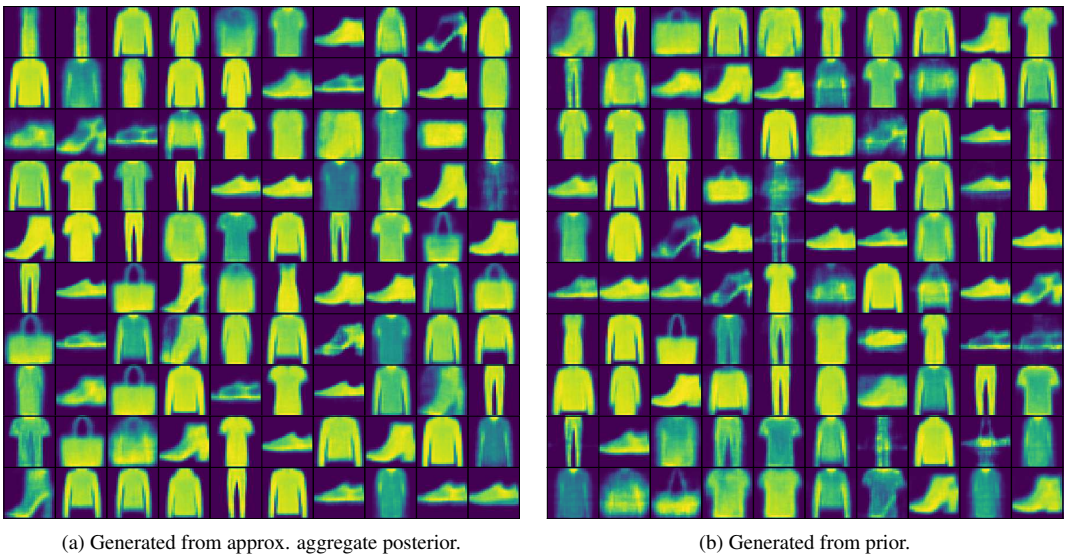

(a) Generated from approx. aggregate posterior.      (b) Generated from prior.

*Figure 9.* 100 Fashion MNIST images generated by DIP-VAE where a lower bound of the ELBO loss is being optimised.

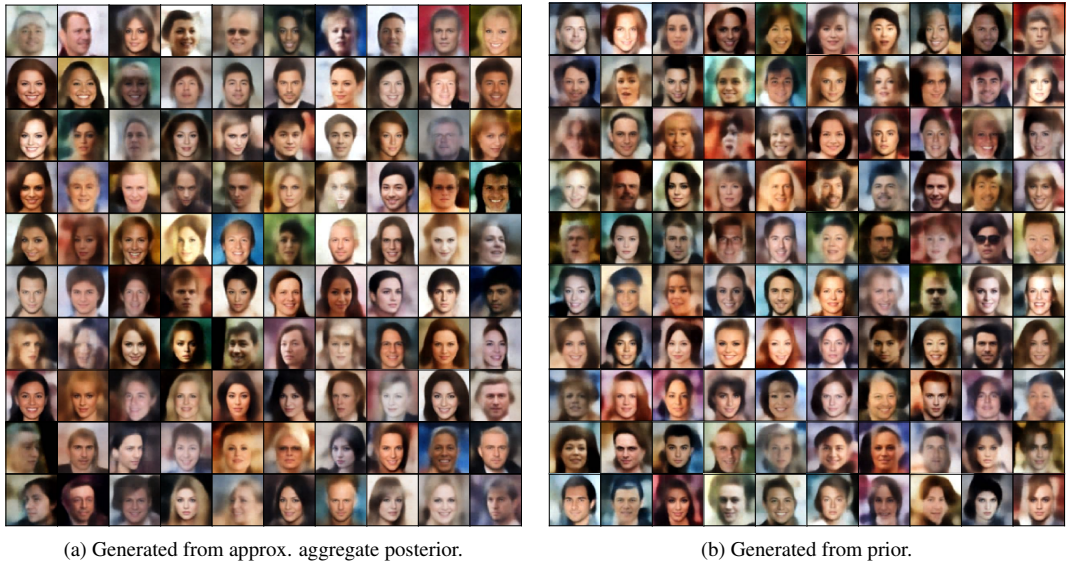

(a) Generated from approx. aggregate posterior.

(b) Generated from prior.

*Figure 10.* Generated images given by our proposal where the variance in the likelihood model is learnt. Some corrupted samples appear occasionally in (a). We think this is due to the approximation error of the true aggregate posterior.

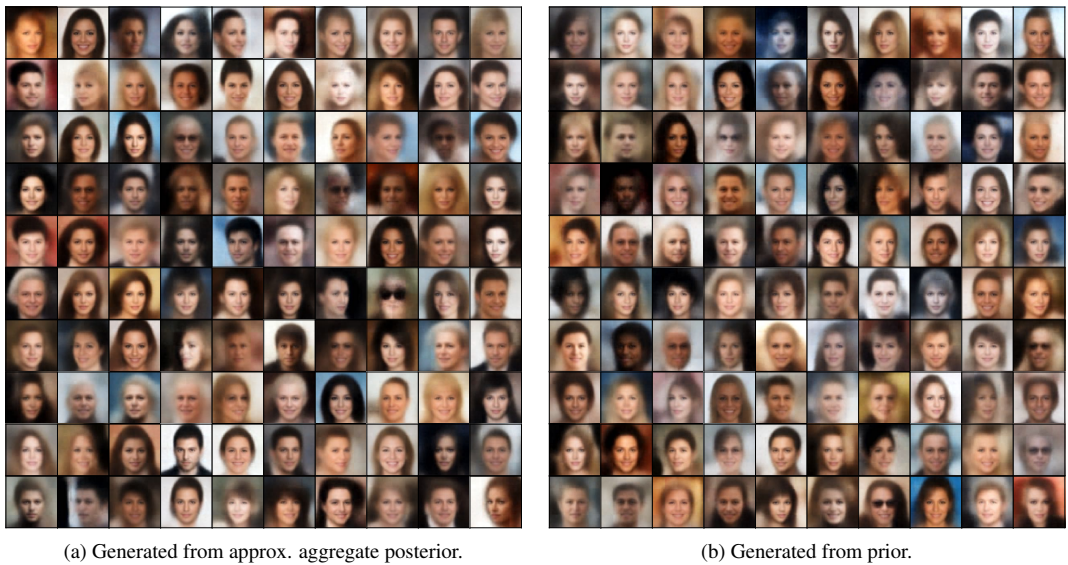

(a) Generated from approx. aggregate posterior.

(b) Generated from prior.

*Figure 11.* Generated images given by the VAE learning method where the variance in the likelihood model is fixed at 0.5.

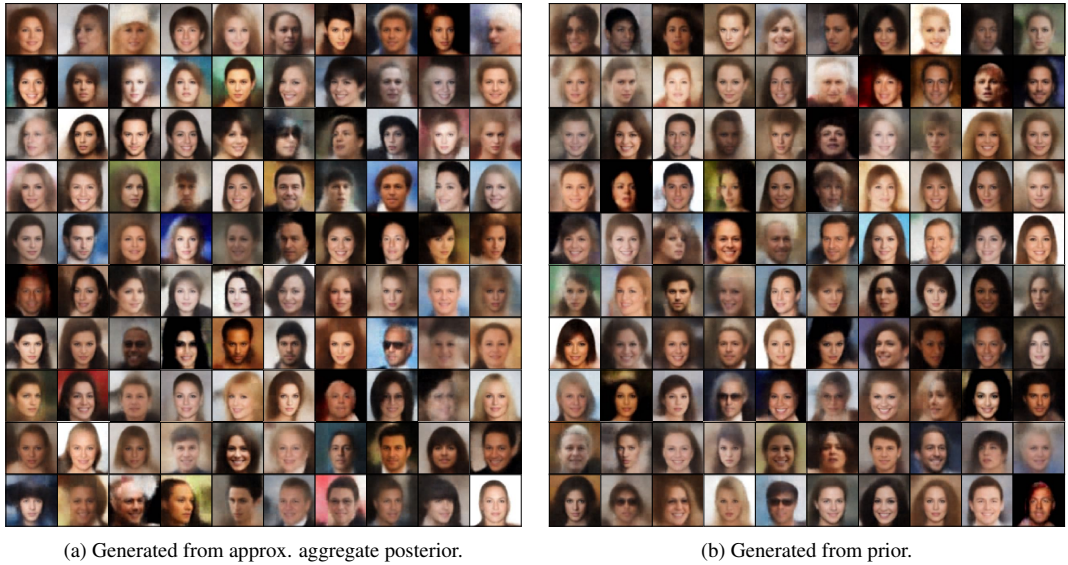

(a) Generated from approx. aggregate posterior.
(b) Generated from prior.

*Figure 12.* Generated images given by the $\beta$-VAE learning method where the variance in the likelihood model is fixed at 1.

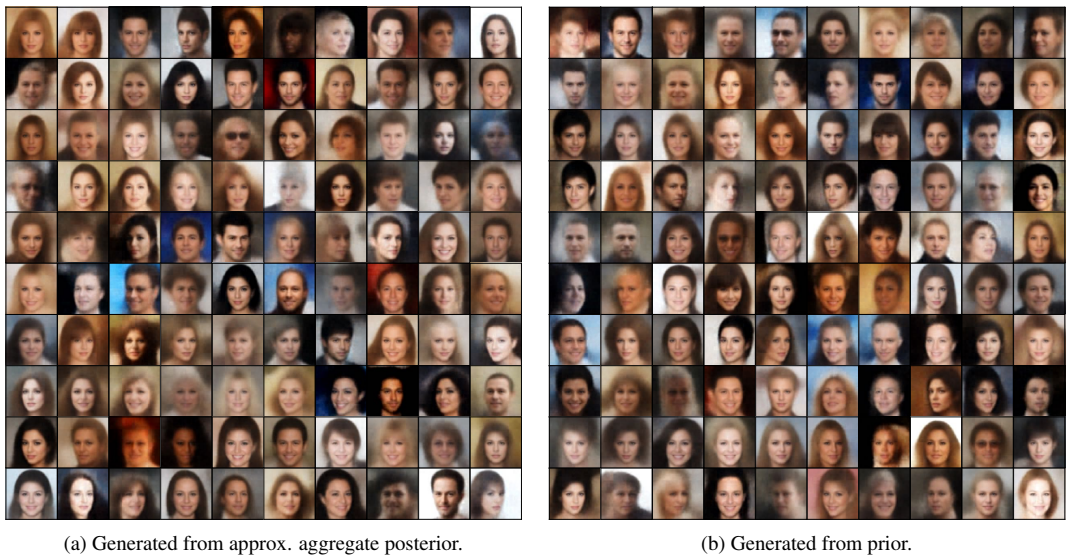

(a) Generated from approx. aggregate posterior.
(b) Generated from prior.

*Figure 13.* Generated images given by the DIP learning method where a lower bound of the ELBO loss is being optimised.

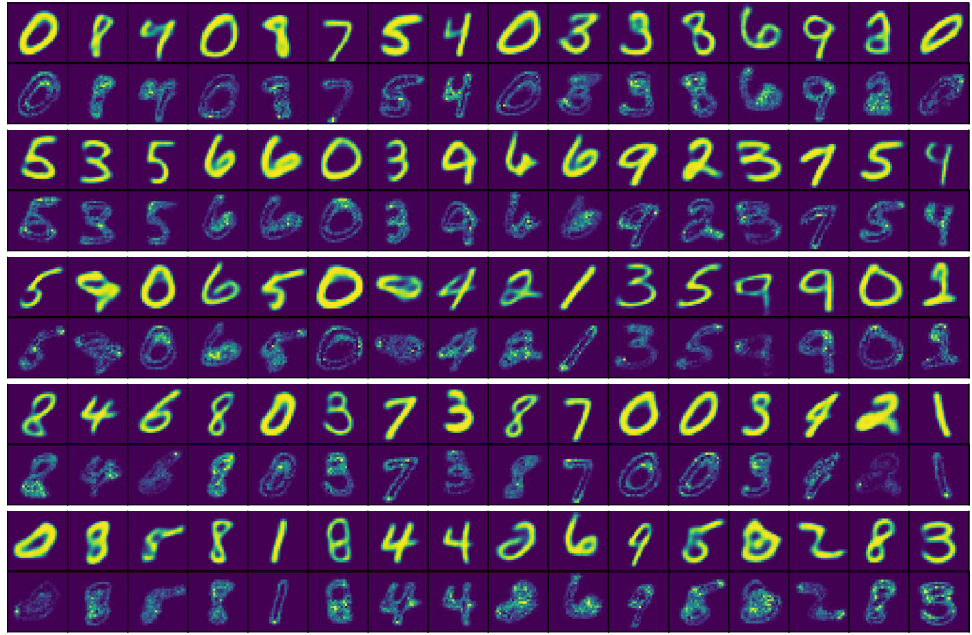

Figure 14. Generated samples for MNIST dataset given by our learning method and the corresponding variance values indicating regions of high uncertainty in the generated samples.

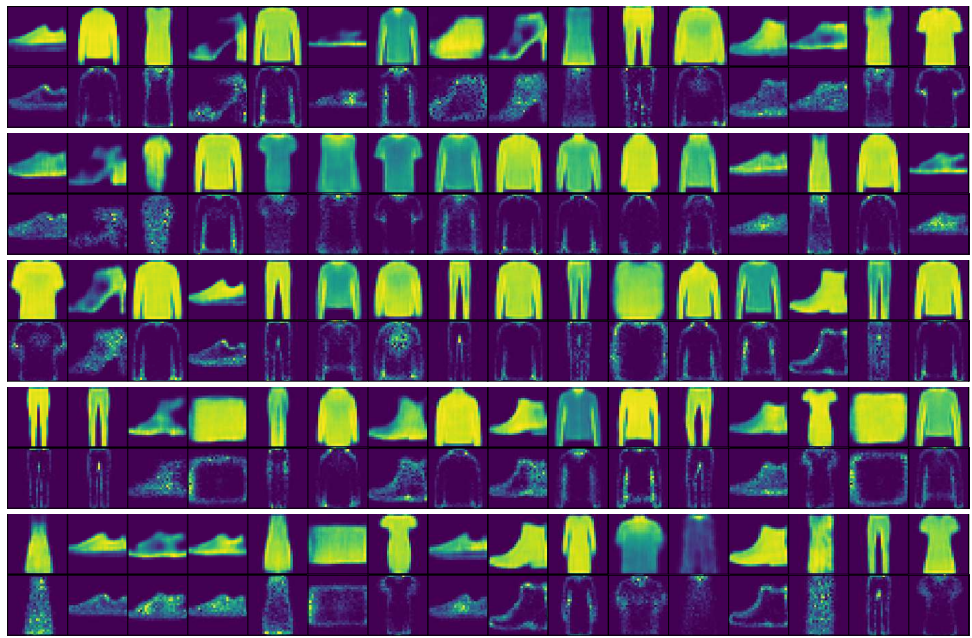

Figure 15. Generated samples for Fashion MNIST dataset given by our learning method and the corresponding variance values indicating regions of high uncertainty in the generated samples.

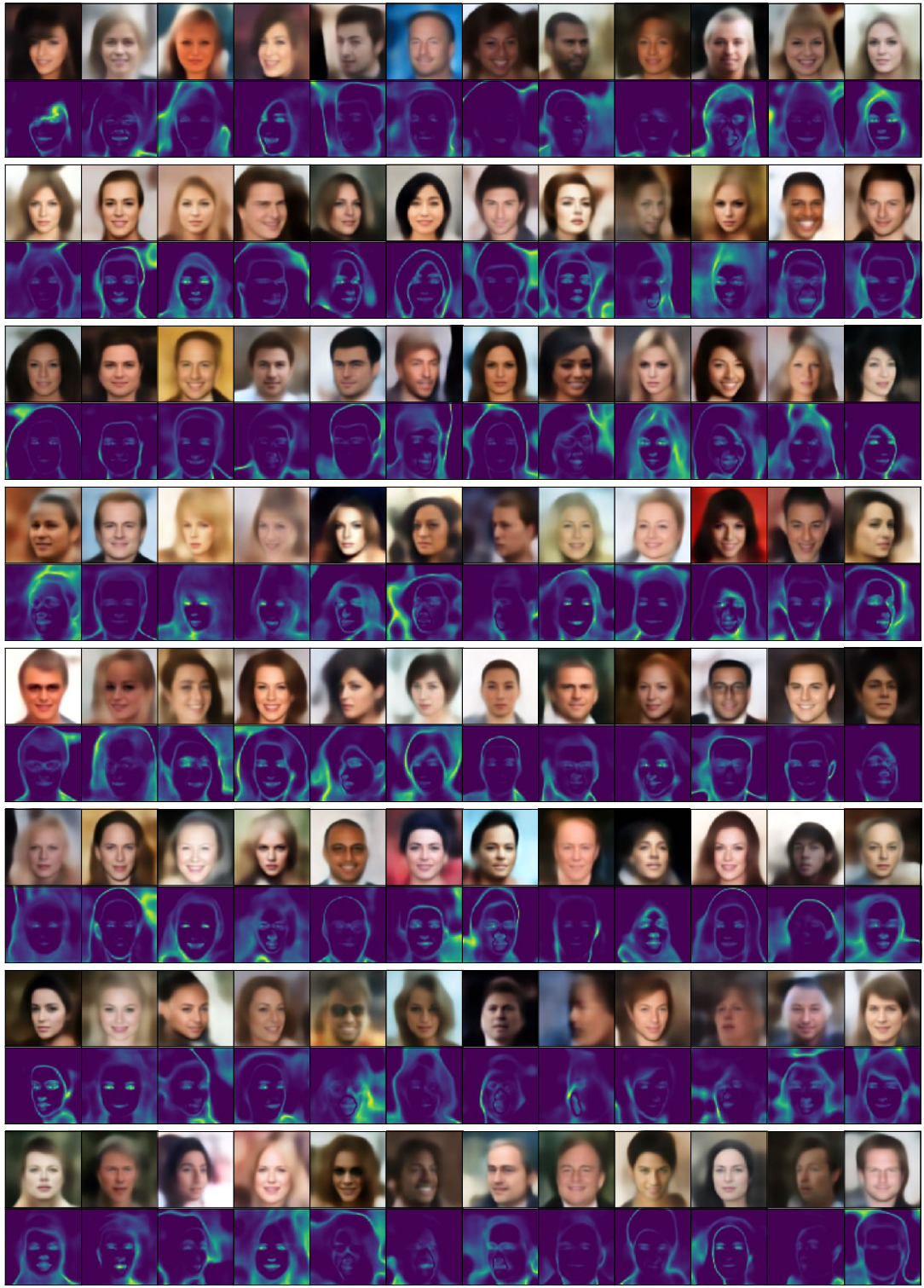

*Figure 16.* Generated samples for CelebA given by our learning method dataset and the corresponding variance values indicating regions of high uncertainty in the generated samples.

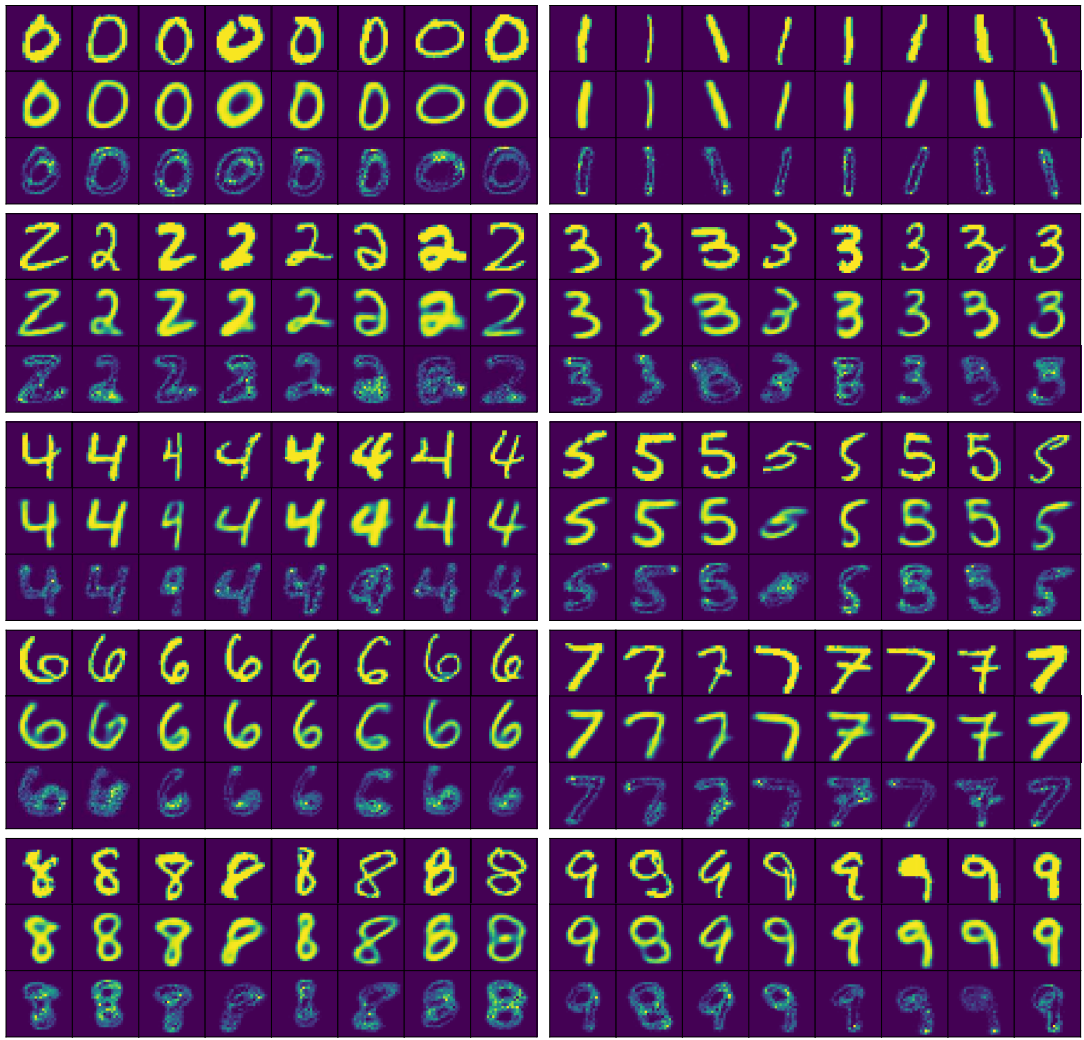

*Figure 17.* Original and reconstructed samples of MNIST dataset (top and middle row for each digit block) and the corresponding variance values (bottom row) indicating regions of high uncertainty in the reconstruction.

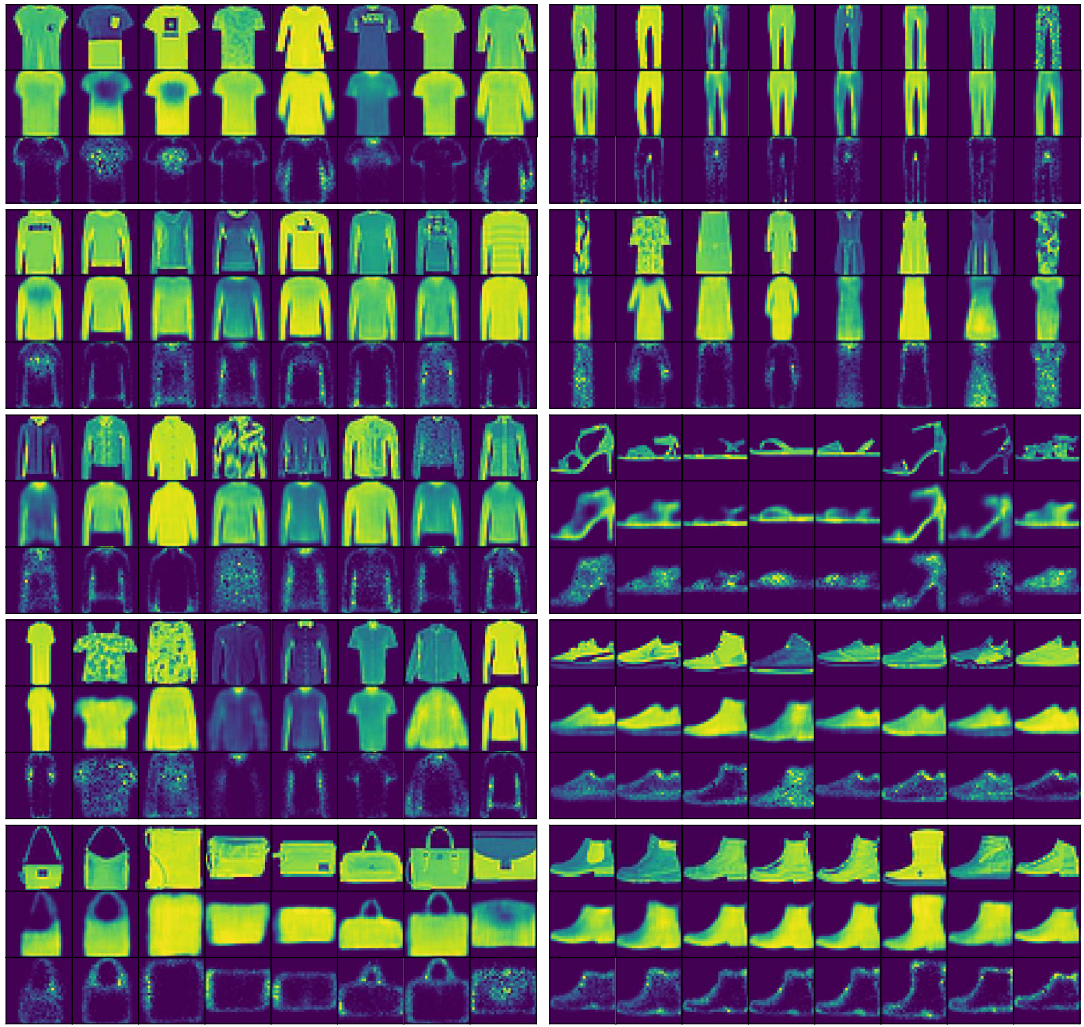

*Figure 18.* Original and reconstructed samples of Fashion MNIST dataset (top and middle row for each digit block) and the corresponding variance values (bottom row) indicating regions of high uncertainty in the reconstruction.

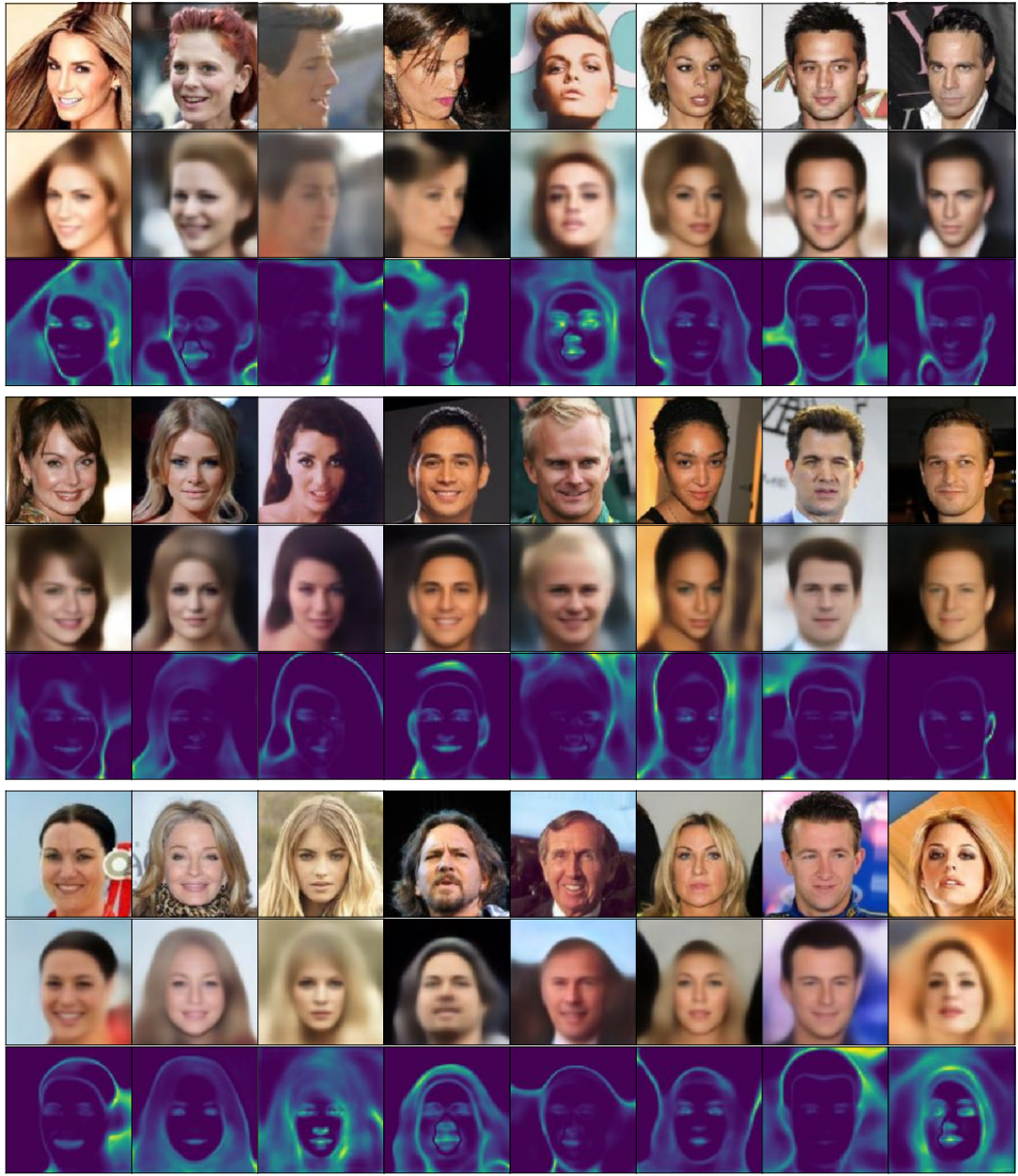

*Figure 19.* Original and reconstructed samples of CelebA dataset (top and middle row for each block) and the corresponding variance values (bottom row) indicating regions of high uncertainty in the reconstruction.



\end{document}